\documentclass[]{fairmeta}

\usepackage[utf8]{inputenc} 
\usepackage[T1]{fontenc}    
\usepackage{url}            
\usepackage{booktabs}       
\usepackage{amsfonts}       
\usepackage{nicefrac}       
\usepackage{microtype}      
\usepackage{xcolor}         
\usepackage{enumitem}
\usepackage{comment}
\usepackage{csquotes}
\usepackage{wrapfig}
\usepackage{afterpage}

\usepackage{subcaption}
\usepackage{rotating}
\usepackage{stackengine}

\usepackage{amsmath}
\usepackage{algorithm}
\usepackage{algpseudocode}
\usepackage{graphicx}
\usepackage{amsthm}
\usepackage{amssymb}
\usepackage{thmtools}
\usepackage{thm-restate}

\definecolor{highlightcolor}{RGB}{255,255,0}

\newtheorem{theorem}{Theorem}

\usepackage[colorinlistoftodos,bordercolor=orange,backgroundcolor=orange!20,linecolor=orange,textsize=scriptsize]{todonotes}

\title{ScheduleFree+: Scaling Learning-Rate-Free \& Schedule-Free Learning to Large Language Models}

\author[1]{%
  Aaron Defazio
}

\affiliation[1]{FAIR at Meta Super-Intelligence Labs}

\abstract{Schedule-Free Learning has shown promise as a practical anytime training method for machine learning, showing success across dozens of standard benchmark problems. However, strong performance for LLM training has only been demonstrated at small scales. We identify a number of fixes necessary to scale up Schedule-Free Learning to larger batch sizes and model sizes, and present a learning-rate-free and schedule-free method (ScheduleFree+) for training large language models which greatly outperforms Warmup-Stable-Decay (WSD) schedules. We also demonstrate that Schedule-Free Learning is most effective for long duration training, and at 1000 tokens per parameter, it outperforms SOTA schedules by 31\%.
Schedule-Free Learning provides a theoretical foundation for the use of model averaging and checkpoint merging during pretraining.}

\date{\today}
\correspondence{\email{adefazio@meta.com}}

\begin{document}

\maketitle

\section{Introduction}

There should be no YOLO training runs. Machine Learning should be stable, smooth and predictable. ML Optimization research, recently laser-focused on building faster training methods, has recently seen a resurgence in methods whose goal is to obtain more reliable convergence with less tuning, all without compromising speed.
\setlength{\intextsep}{5pt}%
\setlength{\columnsep}{5pt}%
\begin{wrapfigure}{r}{0.45\textwidth}
    \centering
    \includegraphics[trim={3mm 3mm 2.5mm 2.5mm},clip,width=0.45\textwidth]{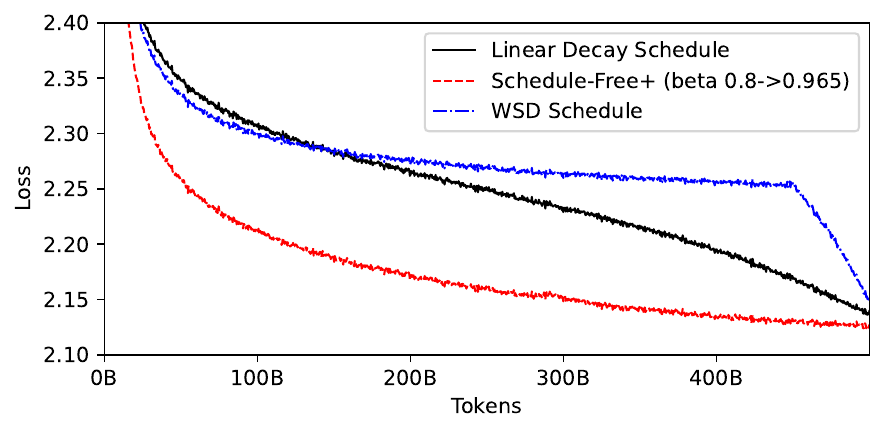}
    \caption{\label{fig:intro}\footnotesize 1000 tokens per parameter run with a 500M parameter model}
    \vspace{-2em}
\end{wrapfigure}

\vspace{-1em}
Schedule-Free Learning \citep{defazio2024roadscheduled} is a recently developed approach that aims to tackle these objectives. Traditional learning-rate schedules are completely replaced by an aggressive form of averaging. Averaging accomplishes the same convergence goal as scheduling, but produces far more stable and smooth loss curves, and gives highly predictable training behavior.

The switch from scheduling to averaging is an enormous change to the optimization process, requiring careful consideration of every other component of the optimizer. This tech report is a deep dive into the application of Schedule-Free Learning to Large Language Models. We identify a number of modifications necessary to match the performance of schedule-based approaches, and we show that Schedule-Free Learning particularly excels at long duration training runs, outperforming schedule-based baselines at 1000 tokens per parameter, giving a \emph{31\% reduction in training time} to reach the same loss. No final loss advantage is seen for short duration (20-100 TPP) runs, although the anytime stopping capability of Schedule-Free Learning is useful at any duration.

\subsection*{Summary of Key Ideas}
\begin{enumerate}
    \item Reintroduction of inner momentum into Schedule-Free training fixes large-batch training divergence.
    \item A proposed variant of the Polyak step size gives a practical learning-rate-free optimizer for language model training.
    \item Warm-starting Schedule-Free further improves early loss values.
    \item Using increasing outer-momentum in Schedule-Free gives large improvements for long training runs.
\end{enumerate}
We call the combination of these approaches ScheduleFree+. We see large improvements in loss compared to WSD Schedules and even compared to state-of-the-art Linear Decay schedules. 

\section{Schedule-Free Learning}
Consider a base optimizer that updates the sequences of iterates $z_t$ using a step direction $G$ computed from the stochastic gradient at $z_t$:
\begin{equation}
z_{t+1} = z_{t} - \gamma_t G_t(z_{t}).
\end{equation}

Here $G$ could be the step direction of AdamW, Muon or SGD for example. Schedule-Free Learning modifies the learning process by introducing an average iterate buffer $x_t$, that computes an online average of $z_t$:
\begin{equation}
x_{t+1} = \frac{t}{t+1}x_{t}+\frac{1}{t+1}z_{t+1},
\end{equation}
so that $x_{T}=\frac{1}{T}\sum_{t=1}^{T}z_{t}$. The idea of maintaining an average point is fundamental to modern optimization theory. Classical convex convergence analysis actually focuses on the average iterate $x$, NOT $z$, however $x$ by itself converges very poorly on neural network training runs. The key enabling idea that takes averaging from unusable to state-of-the-art, is to compute new gradients at a point $y$ that \emph{mixes in} the average, using an interpolation between the averaged and non-averaged points: 
\begin{equation}
y_{t} = \left(1-\beta\right)z_{t}+\beta x_{t}.
\end{equation}
Typically, values of $\beta=0.9$ work well, so that the $y$ sequence is mostly $x$, with a small amount of more recent information from $z$ mixed in. 
The full method is as follows:
\begin{align*}
y_{t} & =\left(1-\beta\right)z_{t}+\beta x_{t},\\
z_{t+1} & =z_{t}-\gamma_t G_t(y_{t}),\\
x_{t+1} & =\left(1-c_{t+1}\right)x_{t}+c_{t+1}z_{t+1},
\end{align*}
This formulation, due to \citet{defazio2024roadscheduled}, includes a more general averaging coefficient $c_{t+1}$, which they set to match $1/t$ eventually, but which provides a more front-weighted average during the learning rate warmup period. The $y$ sequence doesn't need to be stored between steps, it can be computed as needed on each step.

Schedule-Free Learning changes one of the standard aspects of optimizer APIs common to prior methods: the points where you evaluate the gradient, $y$, are no longer your best model estimate. Loss values for the $x$ sequence are much lower, the latest $x_t$ is the model that should be used for evaluations and ultimately returned by the method as output. This can require additional code outside of the usual \texttt{.step()} interface.

The use of averaging means that a decreasing step size is not necessary with Schedule-Free Learning, however a learning rate warmup is still needed for best performance.

\section{Handling Larger Batch-Sizes}

\begin{figure}[htbp]
    \begin{subfigure}{0.49\textwidth}
        \includegraphics[width=\linewidth]{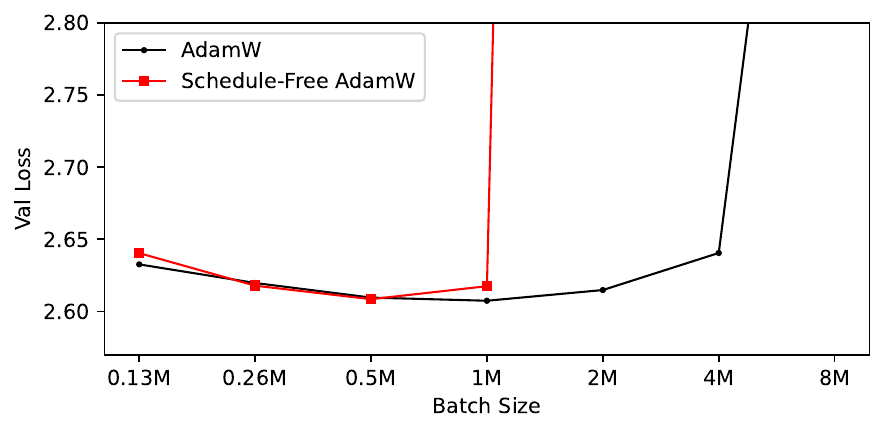}
        \caption{Standard Schedule-Free Learning is not robust to increasing batch-sizes}
        \label{fig:batchsizes-no-mom}
    \end{subfigure}
    \hfill
    \begin{subfigure}{0.49\textwidth}
        \begin{center}
            \includegraphics[width=\textwidth]{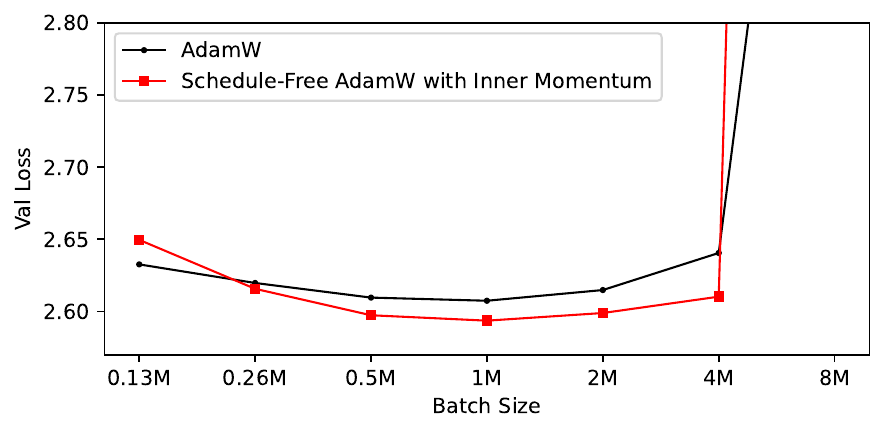}
        \end{center}
        \caption{Inner loop momentum results in greater robustness to larger batch-sizes
        }
        \label{fig:batchsizes-mom}
    \end{subfigure}
    \caption{\label{fig:batchsize}Batch-size scaling on language model pretraining}
    
\end{figure}

Recent evaluations of Schedule-Free Learning on LLM pretraining show that it significantly outperforms Cosine and WSD schedules at medium batch-sizes. However, as the batch-size is increased it falls behind \citep{zhang2024does, morwani2025connections}. This behavior was not identified in \citet{defazio2024roadscheduled} as training was performed on V100 GPUs which can only handle relatively small batch-sizes.

In Figure~\ref{fig:batchsize}, we investigate the behavior of Schedule-Free AdamW against AdamW with a Linear-Decay Schedule \citep{defazio2023when} as the batch-size is varied. This task uses a 120M parameter Llama 3 architecture transformer trained on the FineWeb-EDU dataset for 8B tokens. Warm-starting from the same 2B token checkpoint is used for all runs, which provides better performance for large-batch sizes. For both methods, the learning rate was tuned at batch-size 1M, and set using a $\sqrt{b}$ scaling law for other batch-sizes. This square-root scaling law for learning rate given batch-size is well-established, and we confirmed experimentally that it outperforms linear scaling on this problem. This scaling is not perfect, and for the smallest batch-size we see worse loss values for Schedule-Free than can be obtained by manually tuning at that batch-size.

Figure~\ref{fig:batchsizes-no-mom} shows that Schedule-Free falls behind in terms of loss for larger batch-sizes, and hits a scaling ``cliff'' at 2M tokens per batch, compared to AdamW which has a similar cliff at 4M tokens.

This behavior can be fixed by \textbf{reintroducing inner momentum} into Schedule-Free Learning. \citet{defazio2024roadscheduled} removed the momentum from AdamW when applying Schedule-Free Learning on top of it as it did not provide any benefit in their early experimentation on small models, and by removing inner momentum they were able to reduce the memory usage of Schedule-Free Learning. 

Figure~\ref{fig:batchsizes-mom} shows that Schedule-Free AdamW with inner momentum (here $\beta_1=0.75$) performs better at small batch-sizes, while also allowing the use of batch-sizes of the same magnitude as regular AdamW. 

\subsection*{Why does inner momentum help with larger batch-sizes?}
Momentum has been shown experimentally to help with large-batch training, while providing little or no improvement at small batch-sizes. \citet{shallue2019measuring} show this effect clearly for vision models, and \citet{marek2025small} is a more recent comparison using a LLM pretraining task, showing that this phenomenon is surprisingly universal. \citet{zhang2019algorithmic} established that exactly this behavior can be proven to hold under a noisy-quadratic model of the objective function. 

We believe this behavior is due to enhanced robustness to larger step-sizes, rather than to larger batch-sizes, as momentum allows stable training under larger step-sizes even for smaller-batch sizes. As the batch-size is increased, the optimal learning rate increases, and that optimal learning rate becomes unstable earlier when no momentum is used. This is shown in Figure~\ref{fig:bs64-lr} for training at batch-size 0.5M with traditional Schedule-Free ($\beta_1=0$) and with Inner momentum ($\beta_1=0.9$). Training is stable and achieves the same loss for smaller learning rates, however the zero-momentum version starts to have worse loss and then diverges as higher learning rates are used.

\subsection*{Investigating C-Refinement for large batch training}
\citet{song2025through} propose a fix for large-batch training that modifies the averaging weights. They propose modifying the averaging weights for $x$ in the update $x_{t+1} = \left(1-c_{t+1}\right)x_{t} + c_{t+1}z_{t+1}$ to:
\begin{equation}
c_{t+1}=\frac{(1-\beta)C}{t+1}.
\end{equation}
\begin{figure}[t]
    \centering
    \begin{subfigure}{0.49\textwidth}
    \includegraphics[width=\linewidth]{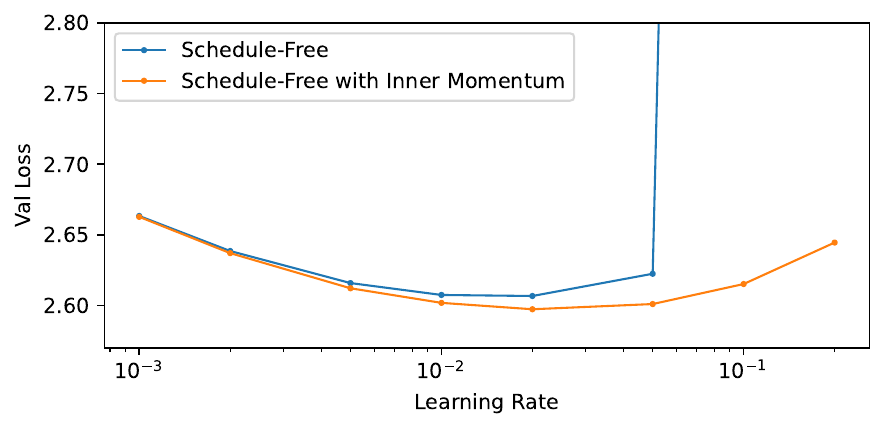}
    \caption{Inner momentum of $\beta_1=0.9$ enables training with larger learning rates}
    \label{fig:bs64-lr}
    \end{subfigure}
    \begin{subfigure}{0.49\textwidth}
    \includegraphics[width=\linewidth]{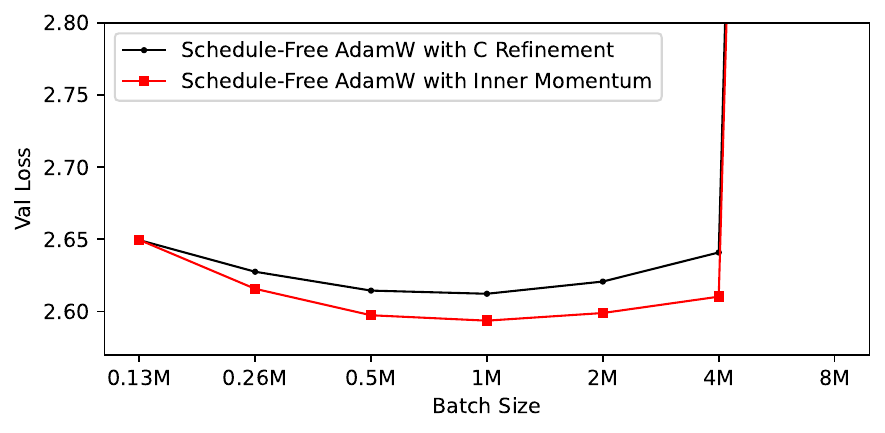}
    \caption{Using inner momentum gives lower loss at larger batch sizes than using C-Refinement}
    \label{fig:bs-refine}
    \end{subfigure}
    \caption{\label{fig:inner-mom}Inner momentum comparison}
\end{figure}
Where C is a new hyper-parameter constant. When rewriting the update of $x$ purely in terms of $y$ (i.e. eliminating $z$), this version removes the dependence of the averaging weights for $x$ on $\beta$. \citet{defazio2024roadscheduled} discuss a tunable hyper-parameter $r$ which can also be used to control the weighting sequence, via the update:
\begin{equation}
    c_{t+1} = \frac{r+1}{t+r+1}.
\end{equation}
The $C$ hyper-parameter can be seen as performing a similar function to $r$ but parameterized differently. The reference implementation uses a different, approximate implementation of $r$-weighting, which we discuss further in Section~\ref{sec:annealing-and-weighting}. We implemented the $C$ version as given by \citet{song2025through} and tested it for large batch-training at $C=50$ and $C=200$. For this test problem, $C=50$ performed slightly better. Figure~\ref{fig:bs-refine} shows that using C-Refinement does enable training at larger batch sizes, but is not as effective as our inner momentum approach.

\section{Handling Gradient Norm Drift}

\begin{figure}[t]
    \begin{subfigure}{0.49\textwidth}
        \includegraphics[width=\linewidth]{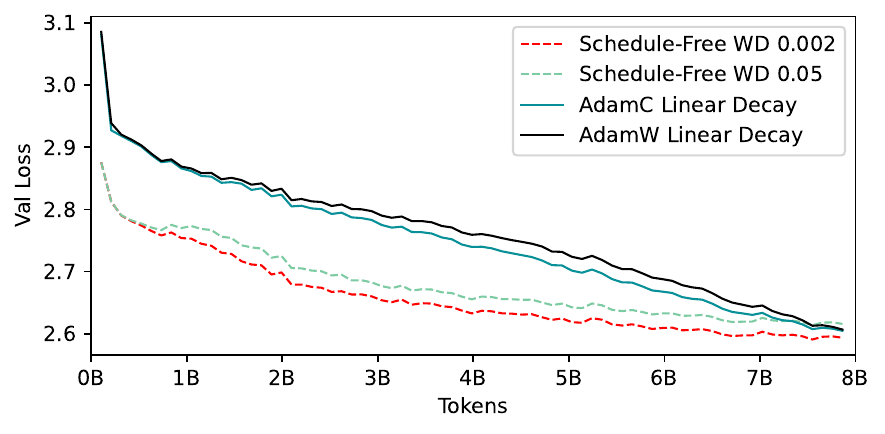}
        \caption{Larger "normal" levels of weight decay are harmful for Schedule-Free Learning}
        \label{fig:wd-loss}
    \end{subfigure}
    \hfill
    \begin{subfigure}{0.49\textwidth}
        \includegraphics[width=\textwidth]{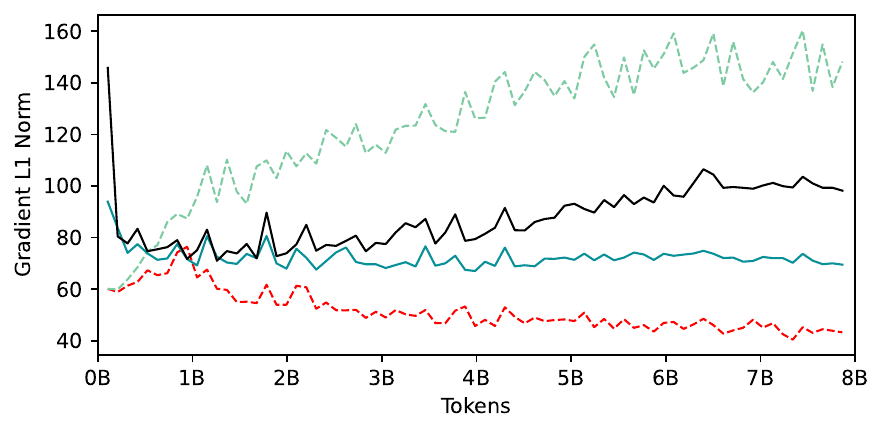}
        \caption{Unexpectedly, gradient norms are flatter for smaller WD values for Schedule-Free Learning}
        \label{fig:wd-gnorms}
    \end{subfigure}
    \caption{The effect of weight-decay on gradient norms}
    \label{fig:wd}
\end{figure}
\begin{figure}[t]
    \centering\begin{subfigure}{0.49\textwidth}
        \includegraphics[height=8cm]{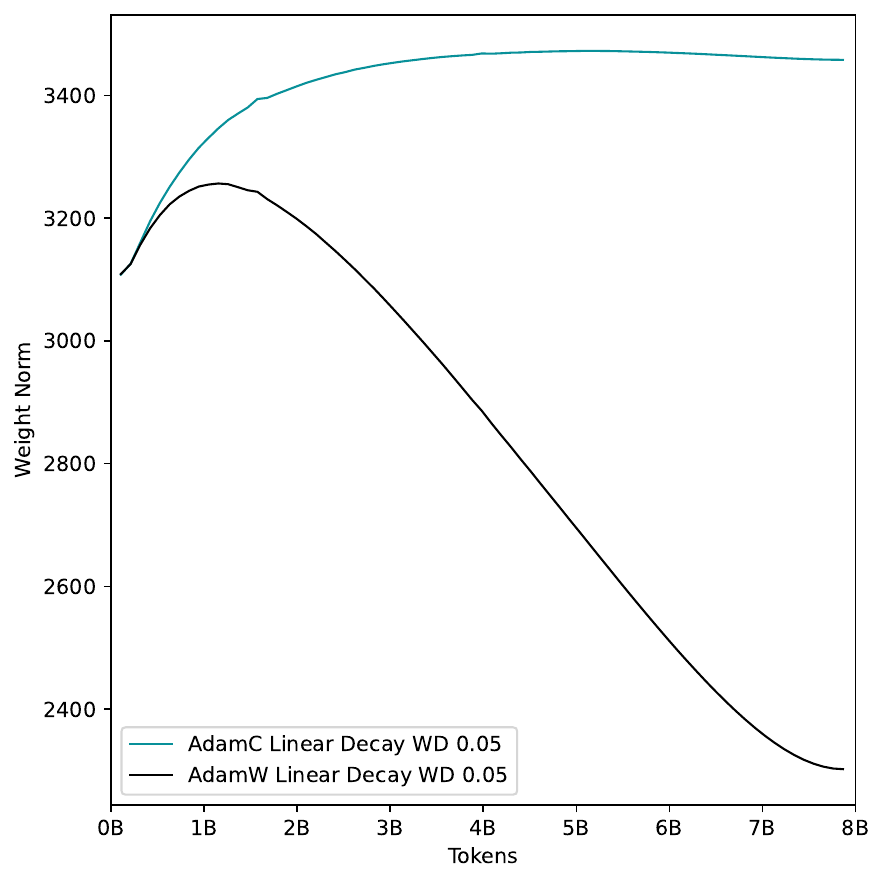}
        \caption{Linear Decay Schedules with WD 0.05: AdamC produces more stable norms}
        \label{fig:avg-norm}
    \end{subfigure}
    \begin{subfigure}{0.49\textwidth}
        \includegraphics[height=8cm]{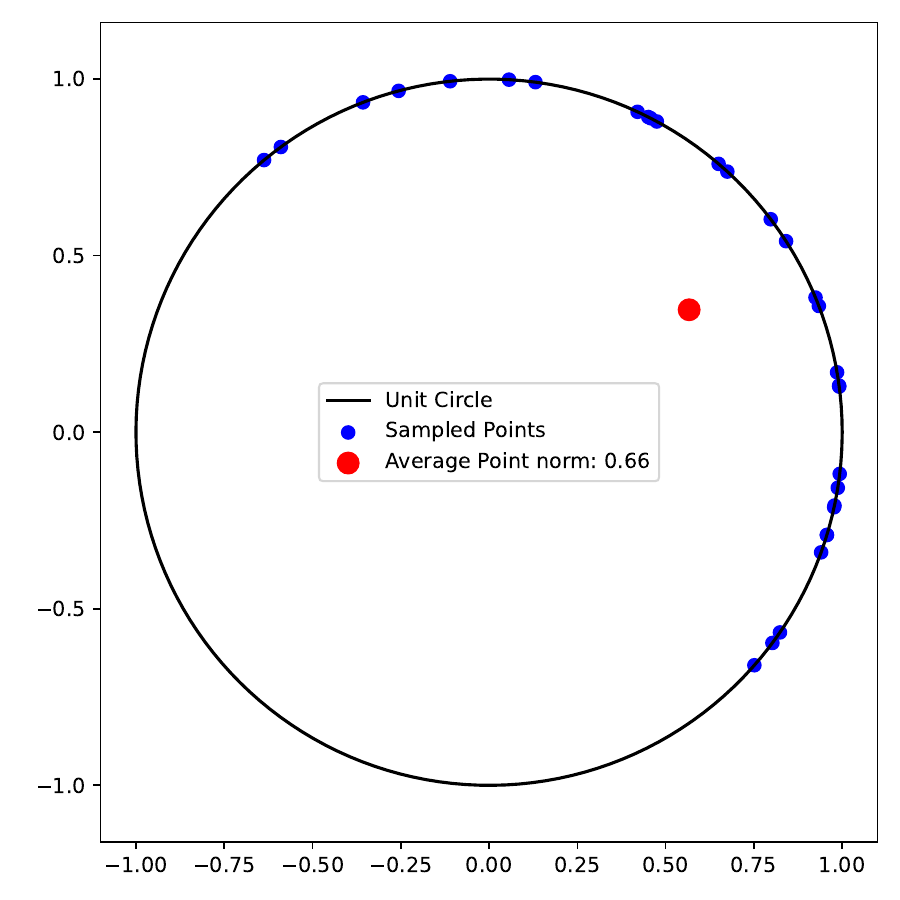}
        \caption{Averages have lower weight norms when weights are on the unit sphere}
        \label{fig:wd-ld-norm}
    \end{subfigure}
    
    \begin{subfigure}{0.49\textwidth}
        \includegraphics[width=\textwidth]{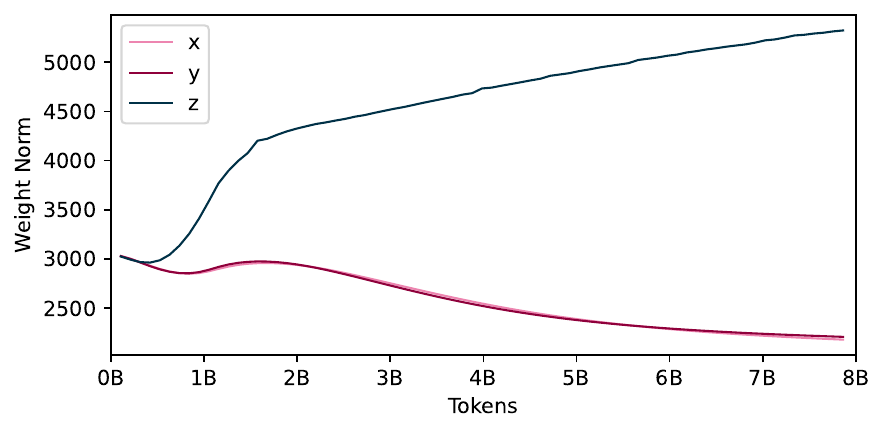}
        \caption{Schedule-Free with WD 0.05:  Weight norms decrease for the x/y sequences}
        \label{fig:wd-sf05-weight}
    \end{subfigure}
    \begin{subfigure}{0.49\textwidth}
        \includegraphics[width=\textwidth]{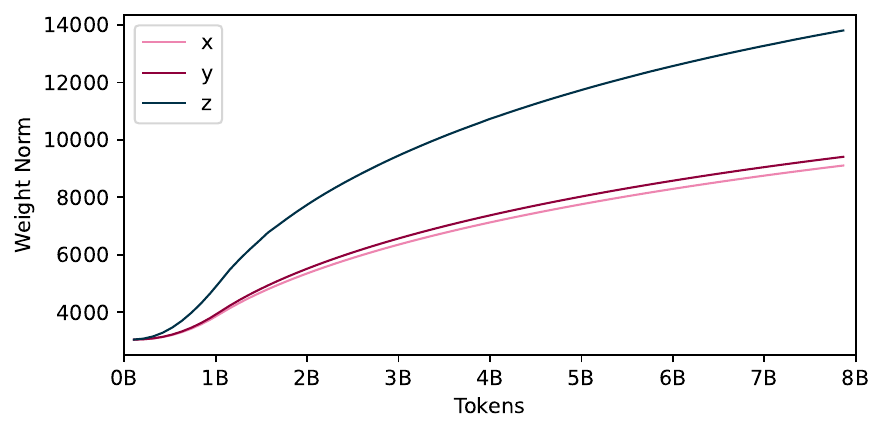}
        \caption{Schedule-Free with WD 0.002 has faster increasing weight norms}
        \label{fig:wd-sf002-weight}
    \end{subfigure}
    \caption{Schedule-Free Learning induces different weight norm dynamics than traditional schedules}
    \label{fig:wd-weights}
\end{figure}


In \citet{defazio2024roadscheduled}, best results are only obtained with unusual values of the weight-decay hyper-parameter. The weight-decay used is much smaller than typical for AdamW. While these settings work, they induce undesirably large weight norms in trained models, particularly for long training runs. In this section, we show that these smaller weight-decay values are actually compensating for gradient norm drift, in a somewhat hacky way for which better solutions exist.

The key clue to understanding this odd behavior lies in the change that occurs in the gradient norms during training. Figure~\ref{fig:wd-gnorms} shows that Schedule-Free Learning with weight-decay $0.002$ produces a more stable gradient norm sequence over time than weight-decay $0.05$, which more than doubles during the course of training. In comparison, regular AdamW with a Linear Decay schedule only shows a $25\%$ change in gradient norm (outside of the initial warmup period), and AdamC, a small modification to AdamW designed to produce flatter gradient norm sequences \citep{defazio2025gradientsrapidlyincreasenear}, has nearly constant gradient norm. 

Constant or near-constant gradient norms during training appears to be empirically beneficial during training. \citet{defazio2025gradientsrapidlyincreasenear} show that AdamC provides clear benefits, particularly for longer training runs. Figure~\ref{fig:wd-loss} shows that AdamC slightly outperforms AdamW here, but the difference is far larger for Schedule-Free Learning, where a large gap in loss is observed between the two Schedule-Free Learning runs.

The underlying reason for this behavior is a simple consequence of the use of normalization layers. \citet{van2017l2} show that networks trained with normalization layers have a different ``effective learning rate'' for each layer that depends inversely on that layer's weight norm. When training with SGD, the effective learning rate is:
\begin{equation}
\eta_{\text{eff}} \propto \eta_{t}/\left\Vert w_{t}\right\Vert ^{2}.
\end{equation}
Here the $\propto$ notation indicates proportional scaling. Gradient norms depend inversely on weight norms \citep{kosson2024rotationalequilibriumweightdecay, defazio2025gradientsrapidlyincreasenear}, i.e. $\left\Vert w_{t}\right\Vert \propto\left\Vert g_{t}\right\Vert ^{-1}$, and so the effective learning rate is proportional to the gradient norm squared:
\begin{equation}
\eta_{\text{eff}} \propto \eta_{t} \left\Vert g_{t}\right\Vert ^{2}, \label{eq:effective-lr}
\end{equation}
and so an increasing gradient norm indicates an increasing effective learning rate. The optimization theory supporting both Schedules and Schedule-Free Learning suggests that a flat or decreasing learning rate (modulo the schedule) is optimal, and so this increasing effective learning rate can be expected to hurt performance. Although this argument is for SGD, similar arguments suggest that the effective learning rate varies as $\eta_{\text{eff}}=\eta_{t}\left\Vert g_{t}\right\Vert_1$ for Adam-like methods. The gradient magnitude estimates used in the denominator of the Adam step are NOT enough to correct the learning rate.

Now, as we observed in Figure~\ref{fig:wd-gnorms}, Schedule-Free Learning with small weight decay values appears empirically to give flatter gradient-norm sequences. This behavior appears to be induced by growing weight-norms counter-acting the gradient-norm increase. Figure~\ref{fig:wd-sf05-weight} shows that larger WD values induce decreasing weight norms in the $x,y$ sequences used in Schedule-Free Learning, whereas smaller WD values (Figure~\ref{fig:wd-sf002-weight}) give increasing weight norms. We find that when tuning the method to give the best loss, the weight-decay values inducing approximately flat gradient norm sequences give the best loss values. 

Rapidly increasing weight norms are undesirable as they increase instability and cause issues with weight quantization. We found unpredictable and time-horizon dependent behavior when trying to scale to large model sizes and training durations when using small weight-decay values. Instead, flat weight-norms, induced by using larger weight decay values, result in more predictable behavior, but require special handling when using Schedule-Free Learning.

When using flat or near-flat learning rate sequences, weight-decay induces weight norms to lie on a hyper-sphere \citep{defazio2025gradientsrapidlyincreasenear}, overriding values from initialization \citep{kosson2025weightdecaymattermup}. However, when using iterate averages as used in Schedule-Free Learning, the average point will have a smaller norm. This is illustrated schematically in Figure~\ref{fig:wd-ld-norm} in 2D. The average of points on the surface of the sphere will be in the interior of the sphere. This same phenomenon is seen in practice, where the average point's weight norm drops by roughly a factor of 2 as the time progresses before stabilizing (Figure~\ref{fig:l1-wnorms}). 

This is a key problem that must be addressed when applying Schedule-Free Learning: \textbf{Shrinking weight norms induces increasing gradient norms in the model, which affects the effective learning rate} (Equation~\ref{eq:effective-lr}). In the next section we explore the corrections required to fix the convergence of Schedule-Free Learning in this setting.

\section{Inverse-Gradient Norm Weighting is Highly Beneficial}
\begin{figure}[t]
    \begin{subfigure}{0.49\textwidth}
        \includegraphics[width=\linewidth]{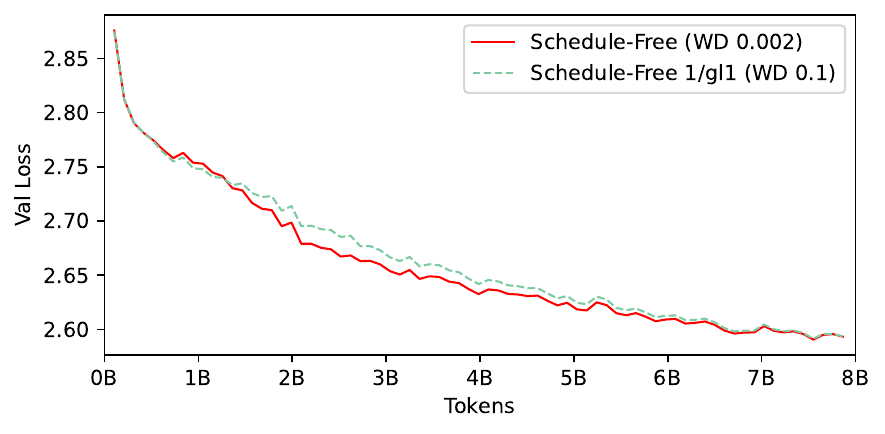}
        \caption{Loss values with and without inverse-gradient norm weighting.\label{fig:l1-loss}}
    \end{subfigure}
    \hfill
    \begin{subfigure}{0.49\textwidth}
        \includegraphics[width=\textwidth]{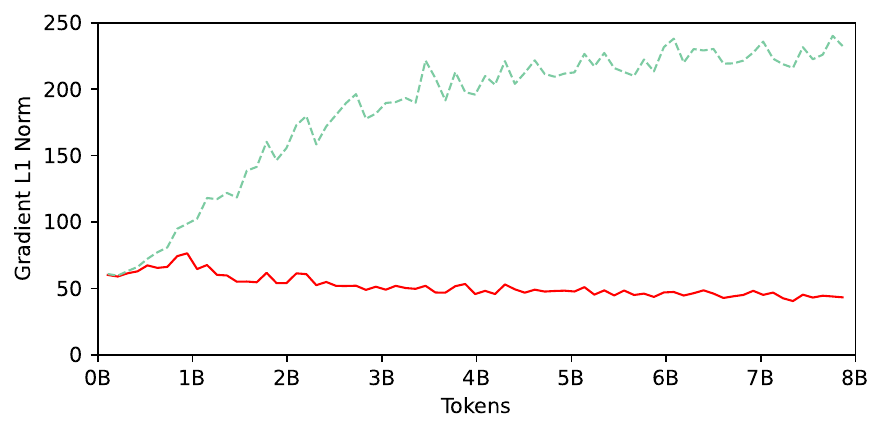}
        \caption{Gradient Norms differ substantially,  rising higher then stabilizing when larger WD values are used.\label{fig:l1-gnorms}}
    \end{subfigure}

    \centering\begin{subfigure}{0.49\textwidth}
    \includegraphics[width=\linewidth]{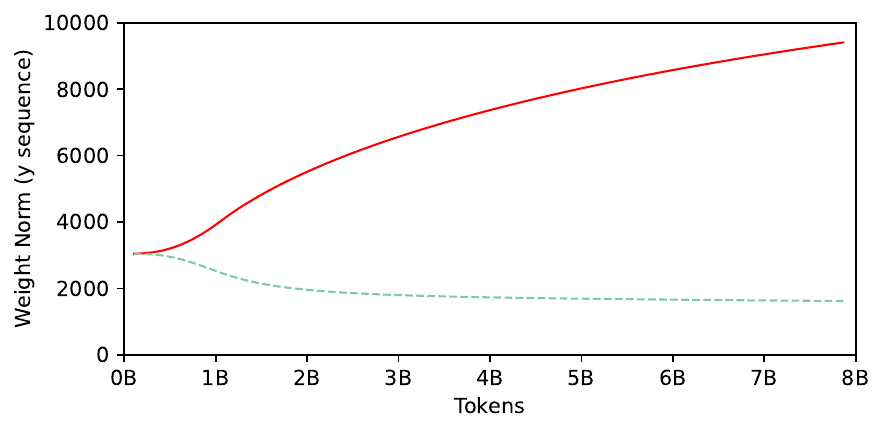}
        \caption{Weight norms for Schedule-Free with larger WD values stabilize after an initial burn-in period.\label{fig:l1-wnorms}}
    \end{subfigure}
    \caption{\label{fig:l1}Inverse gradient L1 norm step sizes give equivalent final loss to using small WD values, while having a stable weight-norm during training}
    
\end{figure}

The theory behind Schedule-Free Learning provides a (anytime) last-iterate convergence guarantee for the $x_T$ sequence that depends on the weighted regret of the base optimizer. In practice, the weights used are typically just the learning rates $\gamma_t$, and the bound is as follows:
\begin{align}
    \mathbb{E}[F(x_T) - F(x_\star)]\le \frac{\mathbb{E}[\sum_{t=1}^T \gamma_t\langle g_t, z_t -x_\star\rangle]}{\sum_{i=1}^T \gamma_i}.
\end{align}
Here $g_t$ is the (sub-)gradient at time-step $t$, and $x_*$ is any minimizer of the (convex) objective function $F$.
As an example, we can apply this to the SGD step $x_{t+1} = x_t - \gamma_t g_t$, which has $\gamma$-weighted regret of \citep{zinkevich2003online}:
    \begin{align*}
        \sum_{t=1}^T \gamma_t \langle g_t,z_t -x_*\rangle& \leq \frac{1}{2}D^2 + \frac{1}{2} \sum_{t=1}^T \gamma_t^2 \|g_t\|^2.
    \end{align*}
The benefit of a bound that depends on weighted-regret rather than standard unweighted-regret is that we have additional flexibility to optimize the right hand side of the bound by optimizing the weights in a problem dependent way, with the caveat that the Schedule-Free theory requires the weight at time $t$ to be statistically independent of the sampling noise at time $t$. The optimal weights, if the gradient-norm sequence is fixed in advance, are given by the following theorem: 
\begin{theorem}\label{thm:optimized-weights} \citep{defazio2023when}
For a fixed sequence $\|g_1\|^2,\dots,\|g_T\|^2$, the value of $\frac{1}{2\cdot \sum_{t=1}^T \gamma_t}(D^2 + \sum_{t=1}^T \gamma_t^2 \|g_t\|^2)$ is minimized by setting $\gamma_t$ proportional to $\|g_t\|^{-2}$.
\end{theorem}
We may approximate this ideal weighting while dodging issues of statistical-independence by using an estimate of the gradient norm that is temporally smoothed, which will still capture the overall scale of the gradient-norm. An exponential-moving average (EMA) of the gradient norm, such as used for a momentum buffer (with dampening), provides such an estimate.

Notice that this result suggests a correction to the learning rate that counter-acts the gradient-norm-squared drift in the learning rate produced by Normalized layers (Equation~\ref{eq:effective-lr}). Although the behavior of normalized layers is non-convex in nature, the above analysis in the convex case suggests the exact same correction! 

\begin{figure}[t]
    \centering
    \includegraphics[width=\textwidth]{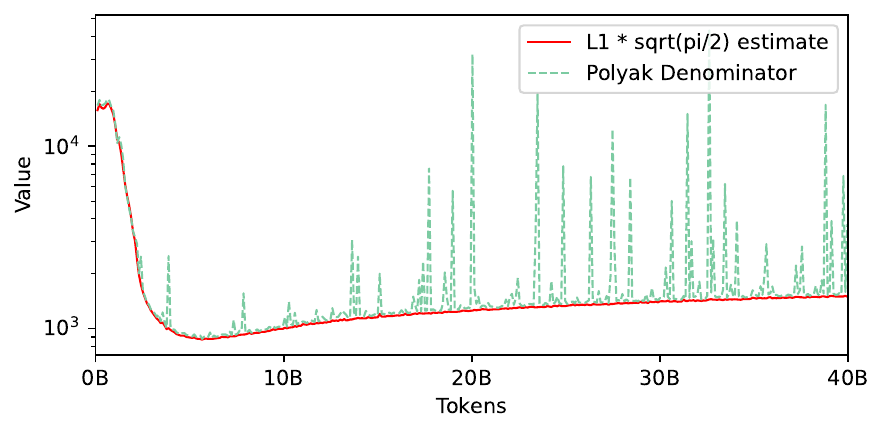}
    \caption{Using the L1 norm approximation to the Polyak denominator gives an accurate and low variance estimate. Training of a 2B model for 40B tokens is shown, with a 0.9 weight EMA used on both sequences. Direct estimation of the Polyak denominator is too noisy to be used in practice.}
    \label{fig:l1_approximation}
\end{figure}

In the case of Adam, the weighting formula changes to:
\begin{equation}
\gamma_{t} \propto \left(\sum_{i=1}^{d}\frac{g_{t,i}^{2}}{\sqrt{v_{t,i}}}\right)^{-1}\approx \left( \sqrt{\frac{\pi}{2}}\left\Vert g_{t}\right\Vert_{1} \right)^{-1}. \label{eq:l1_aprox}
\end{equation}
Here $d$ is the dimension of the vectors, $g_{t,i}$ the $i$th element of $g_t$, and $v$ is the EMA of the square of the gradients maintained in Adam. The $\sqrt{\pi/2}\approx 1.27$ factor comes from converting from squares to absolute values under a normality assumption:
\begin{equation*}
    E[|x|] = \sqrt{\frac{\pi}{2}}\sigma \quad\text{when}\, x \sim \mathcal{N}(0,\sigma^2).
\end{equation*}
This approximation is very accurate in practice and is far more stable than directly using $g$ and $v$ to compute the scaling factor (Figure~\ref{fig:l1_approximation}).

This correction is \emph{highly} effective in practice, as demonstrated in Figure~\ref{fig:l1}. Here we weight by a 0.9 parameter EMA using the inverse (non-squared) gradient L1 norms. The numerator of the learning rate was tuned by grid search. We see that the final validation loss matches our previous best result for the unweighted case, but now we entirely eliminate the issue of rising weight norms, allowing for the training of larger models, and for longer durations without issue.

So we recommend that \textbf{Schedule-Free step-sizes for Adam should be scaled inversely proportional to the L1 norm of the gradient}:
\begin{equation}
\gamma_{t} \propto \frac{1}{\left\Vert g_{t}\right\Vert_{1}}. \label{eq:l1_inverse}
\end{equation}

\section{Learning-Rate Free Learning of LLMs}

Inverse-gradient norm weighting is a key component of the popular Polyak Step size for deterministic optimization:
\begin{equation}
\gamma_{t}=\frac{f(x_{t})-f_{*}}{\left\Vert \nabla f(x_{t})\right\Vert ^{2}}. \label{eq:basic-polyak}
\end{equation}
It's curious that the same inverse-squared gradient weighting appears in the Polyak step as appears in optimal schedules, and the effective learning rate of normalized layers. This suggests that by using the Polyak step size, we can avoid the need to do any additional inverse-gradient norm weighting. It's unclear if there is any deeper connection between these three appearances of the same weighting.

The Polyak step size can be applied to Schedule-Free AdamW training using the approximation in Equation~\ref{eq:l1_aprox} for the denominator:
\begin{equation}
\gamma_{t}=\frac{f(y_{t})-f_{*}+\beta\left\langle \nabla f(y_{t}),z_{t}-x_{t}\right\rangle }{\sqrt{\frac{\pi}{2}}\left\Vert \nabla f(y_{t})\right\Vert _{1}}.
\end{equation}
The numerator also contains a correction factor, which approximates the unknown quantity $f(z_t)-f_*$ which should appear in the numerator, using the Taylor expansion of $f$ around the point $y_t$ as introduced by \citet{oikonomou2026takingroadscheduledadaptive}:
\begin{align*}
f(z_{t}) & \approx f(y_{t})+\left\langle \nabla f(y_{t}),z_{t}-y_{t}\right\rangle \\
 & \approx f(y_{t})+\left\langle \nabla f(y_{t}),z_{t}-\left(\beta x_{t}+(1-\beta)z_{t}\right)\right\rangle \\
 & \approx f(y_{t})+\beta\left\langle \nabla f(y_{t}),z_{t}-x_{t}\right\rangle.
\end{align*}
When applied to stochastic problems, the gradients and function values need to be estimated. We focus on the simplest approach to estimating these quantities: We apply an exponential moving average (coefficient 0.9) to the stochastic estimates of each quantity.

Polyak step sizes have been heavily investigated for their potential in machine learning in the last few years, as they have appealing theoretical properties, however a number of issues arise in their application to stochastic and non-convex optimization problems. In practice, these issues have led to the Polyak step size not being competitive against other adaptive learning rate methods such as Prodigy \citep{mishchenko2023prodigy}. 
The use of Schedule-Free Learning appears to ameliorate each of these issues, resulting in a practical step-size rule.

\begin{figure}[t]
    \begin{subfigure}{0.49\textwidth}
        \includegraphics[width=\linewidth]{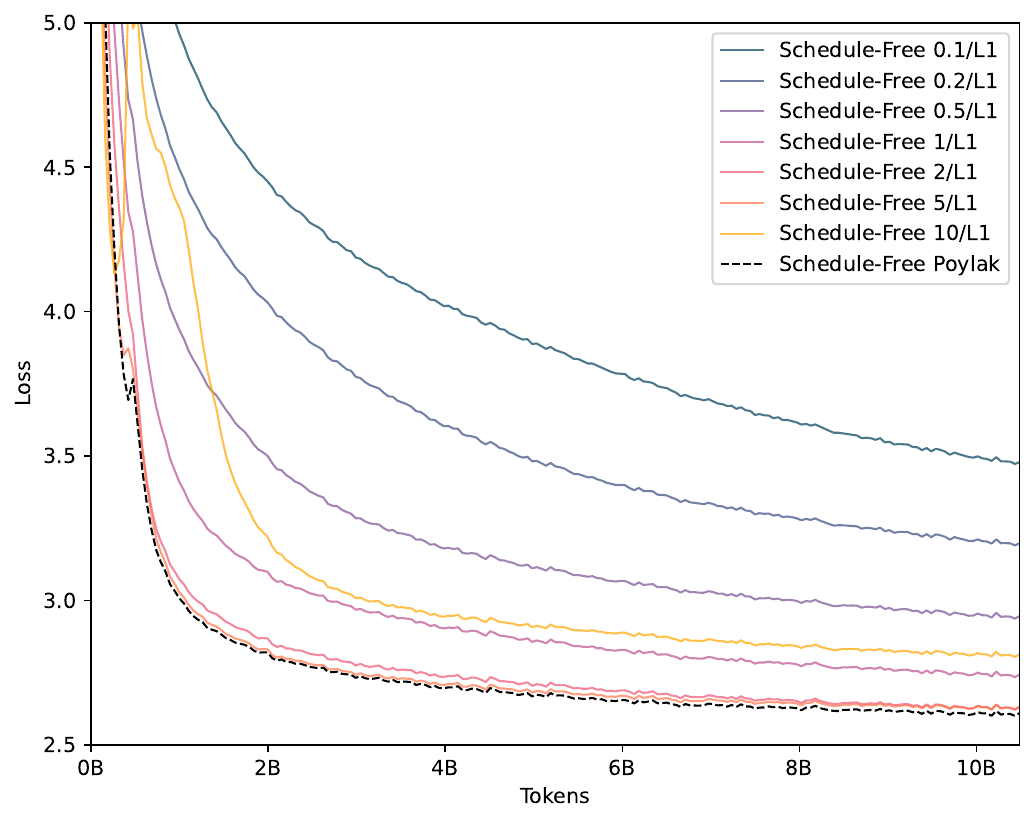}
    \end{subfigure}
    \hfill
    \begin{subfigure}{0.49\textwidth}
        \includegraphics[width=\textwidth]{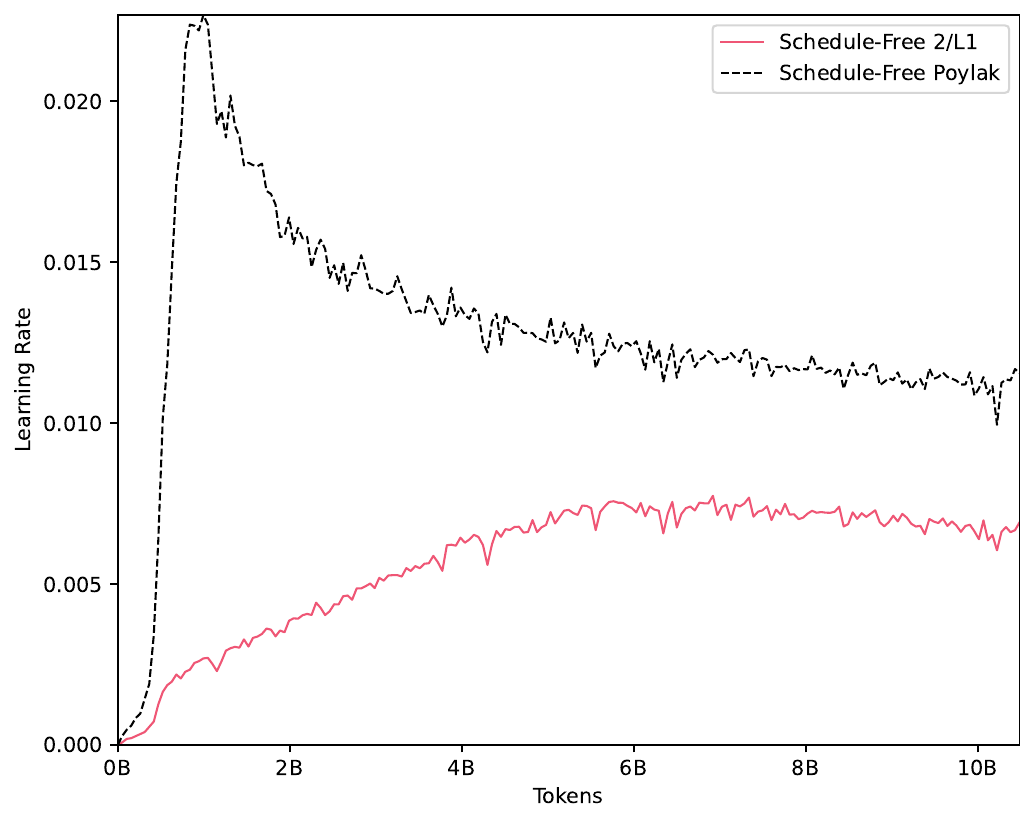}
    \end{subfigure}
    \caption{Polyak step sizes outperform a tuned grid search over L1-adjusted step-sizes}
    \label{fig:polyak}
\end{figure}

\begin{description}
    \item[Vanishing denominators] Small gradients in the denominator result in a blow-up of the step-size, destabilizing training. This can occur simply due to gradient noise during training. Using large-batch sizes greatly help with this issue, but small gradients still  occur when small learning rates are used near the end of training when using a classical learning rate schedule. Schedule-Free avoids this issue by maintaining large learning rates throughout training.
    \item[Sensitivity to $f_*$] When using schedule-based training, the Polyak numerator uses the current iterate $x_t$ in the numerator (Equation~\ref{eq:basic-polyak}). This quantity is noisy, and this noise gets worse as $(x_t)$ approaches $f_*$ during training, giving unstable learning rates. For Schedule-Free Learning, an estimate of $f(z_t)$ is used instead, which is much larger than $f(x_t)$ while having similar noise levels, and therefore lower absolute error. In practice we find that $f(z_t)-f_*$ is a stable quantity to estimate.
\end{description}

The Polyak step size also correctly scales the learning rate as batch-size is varied. When using SGD, the squared-gradient denominator results in a doubling of the learning rate as the batch-size is doubled, following established linear scaling rules. When using AdamW, the learning rate automatically scales proportionally to the square-root of the batch-size, again following standard scaling rules. The Polyak step size also scales the step-size down as training progresses, meaning that no tuning is needed as the number of training tokens is increased.

Figure~\ref{fig:polyak} shows the results of using the Polyak step size with 0.4B tokens of warmup (A linear multiplicative warmup), compared against the $\gamma$/L1 adjusted step sizes with a grid-search of $\gamma$ shown, using the same warmup duration. The Polyak step size outperforms the results of the grid search, and generally uses larger step-sizes than the best performing $\gamma=2$ learning rate. This appears to be due to a higher initial learning rate that comes from the numerator of the step size. \textbf{The Polyak step-size automatically sets the learning rate as effectively as a tuned grid search} and provides the necessary gradient L1-norm adaptivity automatically. 

\subsection{Learning Rate Adaptation Breaks Weight Decay}

In neural networks using normalization layers, weight decay's ($\lambda$) primary effect is to control the eventual magnitude of gradient-to-weight norm ratio \citep{van2017l2}:
\[
\frac{\left\Vert g_t\right\Vert }{\left\Vert z_t\right\Vert }=\sqrt{\frac{2\lambda}{\gamma_t}}.
\]
The gradient/weight ratio can be thought of as a simple dynamical system that rapidly approaches this steady state value. When the learning rate is changing adaptively during training, we end up with a coupled dynamical system, where changes to the learning rate, \emph{feedback into a changing gradient norm}, which then affects the learning rate estimate, and so forth, resulting in erratic, cyclic and potentially divergent behavior. 

This undesirable feedback loop WILL break learning rate adaptation, and we believe it is the primary cause of the failure of adaptive learning rate methods over the last decade. The exception is recently developed methods that only produce non-decreasing learning rate sequences, which can not cycle erratically, such as D-Adaptation \citep{defazio2023dadapt}, Prodigy \citep{mishchenko2023prodigy}, DoG \citep{ivgi2023dog} and DoWG \citep{khaled2023dowg}.

\citet{defazio2025gradientsrapidlyincreasenear} suggests a simple fix for this, modifying the AdamW step to multiply the weight-decay term by an additional \emph{extra} $\gamma/\gamma_{\max}$:
\begin{equation}
z_{t+1}=z_{t}-\gamma_{t}\frac{\hat{m}_{t}}{\sqrt{\hat{v}_{t}}+\epsilon}-\frac{\gamma_{t}^{2}}{\gamma_{\max}}\lambda z_{t}.
\end{equation}
This squared learning rate is counter-intuitive, but a short calculation shows that it results in a steady state independent of $\gamma_t$:
\[
\frac{\left\Vert g_t\right\Vert }{\left\Vert z_t\right\Vert }=\sqrt{\frac{2\lambda}{\gamma_{\max}}}.
\]
When using adaptive learning rates, there is no notion of $\gamma_{\max}$ so we leave out that term, and use \emph{fully-decoupled AdamC}:
\begin{equation}
z_{t+1}=z_{t}-\gamma_{t}\frac{\hat{m}_{t}}{\sqrt{\hat{v}_{t}}+\epsilon}-\gamma_{t}^{2}\lambda z_{t}.
\end{equation}
Giving a steady state of:
\[
\frac{\left\Vert g_t\right\Vert }{\left\Vert z_t\right\Vert }=\sqrt{2\lambda}.
\]
The purpose of the $\gamma_{\max}$ term was to keep the $\lambda$ hyper-parameter on the same scale as regular AdamW. Consequently, Schedule-Free with this modification will require different, larger, weight-decay values, typically in the range 5-50. \textbf{Using fully-decoupled AdamC is necessary when using the Polyak step size.}

\section{Warming-up Schedule-Free}

The original Schedule-Free formulation applied iterate averaging from the beginning of optimization. The weight of each iterate in the average was chosen as proportional to the learning rate squared, which resulted in points during the learning rate warmup phase having lower weight. This approach is very general, however in the case of models where the norm of weights changes dramatically during the early stages of training, this procedure is suboptimal, resulting in slower initial convergence. 

There are several ways to remedy this. Weights can be initialized closer to their final steady state norms. Weight projection can be used, so that weight norms do not change during training. Training can be warm-started from a model trained until weight norms are stable, using a constant learning rate. No fix is needed at all for long training runs. We opted to implement a simple solution, where iterate averaging is not performed for an initial fixed number of steps. This is implemented by setting $c_t=1$ in the Schedule-Free update equations. 

\begin{figure}[t]
    \centering
    \includegraphics[width=0.9\textwidth]{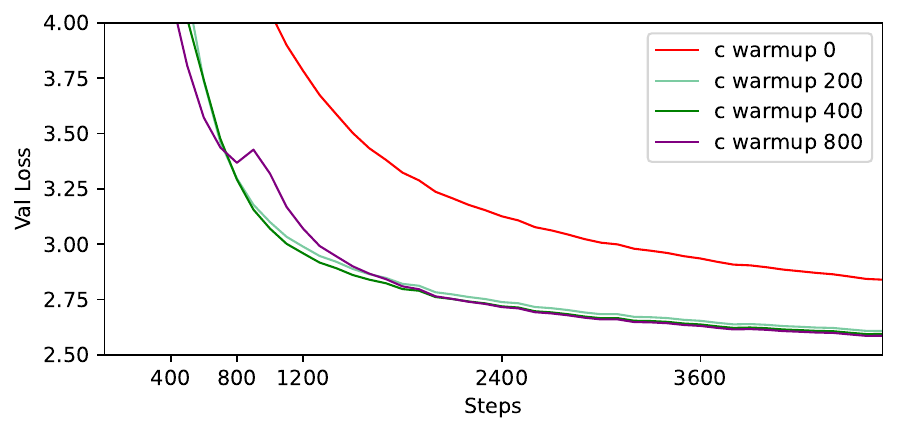}
    \caption{Warm-starting Schedule-Free Learning significantly improves loss values}
    \label{fig:c_warmup}
\end{figure}

Figure~\ref{fig:c_warmup} shows the results of using $c_t$ equal to 1 for $t$ up to 200, 400 and 800 steps, when a learning rate warmup of 400 steps is used. $c$ warmup of 800 steps gave the lowest loss in this example. \textbf{A good heuristic is to do $c$ warmup for 2x the length of the learning rate warmup}. Longer $c$ warmup does introduce a ``hump'' after iterate averaging is introduced, but final loss is significantly lower.

\section{Benefits of Annealing and Weighting}
\label{sec:annealing-and-weighting} The $\beta$ parameter in Schedule-Free Learning interpolates between the raw iterate sequence $z_t$ and its running average $x_t$. This parameter controls a trade-off between the recency of information (the $z_t$ sequence is updated immediately with new gradient information) and the level of smoothing (as an average, the $x_t$ sequence changes slowly). 

\begin{figure}[t]
    \begin{subfigure}{0.49\textwidth}
        \includegraphics[width=\linewidth]{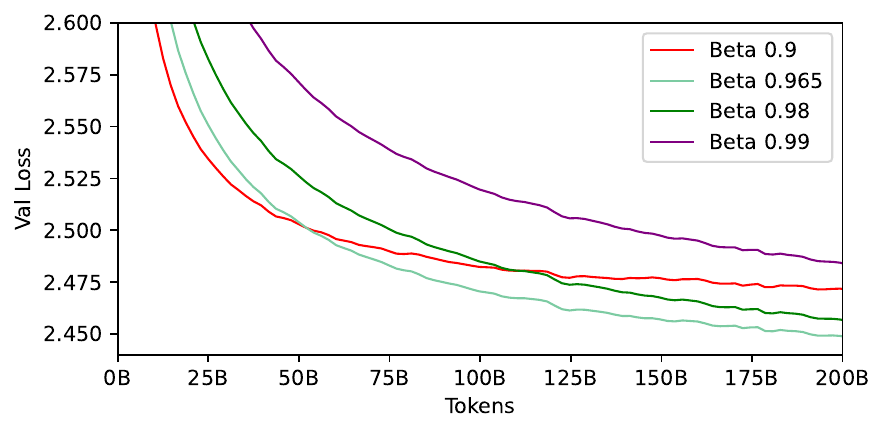}
        \caption{Smaller $\beta$ values are beneficial at the early stages of training, whereas larger values around $0.965$ work better at later stages.}
        \label{fig:sf_beta1-sweep}
    \end{subfigure}
    \hfill
    \begin{subfigure}{0.49\textwidth}
        \includegraphics[width=\textwidth]{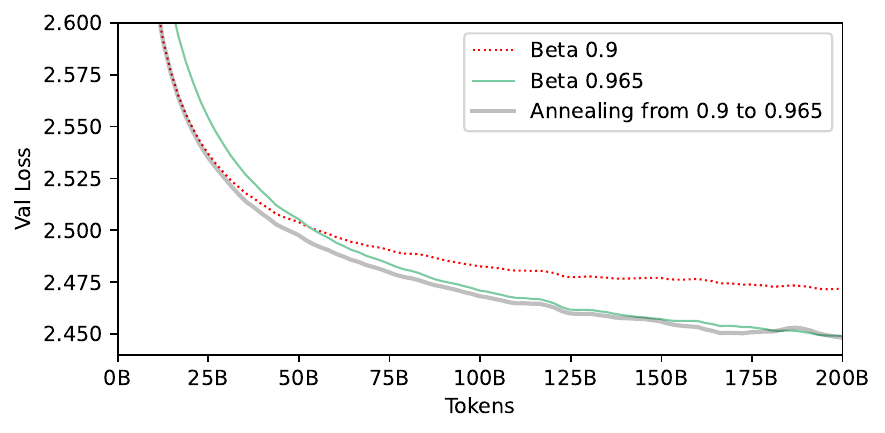}
        \caption{Slowly Annealing $\beta$ during the course of training gives a best-of-both-worlds behavior. Here $\beta$ is log-linearly interpolated from $0.9$ to $0.965$ over the course of the run.}
        \label{fig:sf_beta1-anneal}
    \end{subfigure}
    \caption{Schedule-Free $\beta$ experiments}
    \label{fig:sf_beta1}
\end{figure}

Figure~\ref{fig:sf_beta1-sweep} shows a sweep of the $\beta$ parameter to illustrate this behavior. Low beta values converge faster initially but have worse long run performance. This is another behavior that is not witnessed on smaller scale deep learning problems \citep{defazio2024roadscheduled} but clearly visible for long duration LLM training runs.

Fortunately, we can obtain both rapid early convergence as well as ideal later stage convergence by \emph{annealing} the beta parameter over time. In Figure~\ref{fig:sf_beta1-anneal} we use a time-varying $\beta$ given by a log-linear interpolation:
\begin{equation}
\beta_{t}=1-\exp\left(\left(1-\frac{t}{T}\right)\log\left(1-\beta_{\text{initial}}\right)+\frac{t}{T}\log\left(1-\beta_{\text{final}}\right)\right).
\end{equation}
This results in a smooth interpolation between the two $\beta$ values behavior, capturing the fast initial drop in loss with $\beta_{\text{initial}}$ and the lower later-stage loss matching $\beta_{\text{final}}=0.965$. For long duration training runs, \textbf{annealing Schedule-Free $\beta$ from smaller values like $0.9$ to $0.965$ gives great loss values throughout the training run.}

\begin{figure}[t]
    \centering
    \includegraphics[width=\textwidth]{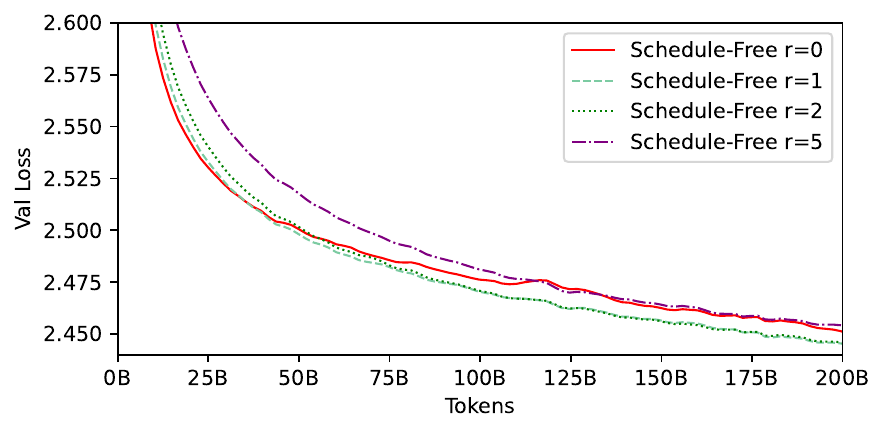}
    \caption{Using values around r=1 is beneficial for longer runs, but slightly degrades performance below 30B tokens.}
    \label{fig:anneal_r}
\end{figure}
The reference implementation of Schedule-Free Learning includes a parameter $r$ that controls the weights of points within the weighted average $x$. The weight at time $t$ is given by:
\begin{equation}
    w_t = t^{r}\max_{i\leq t}\gamma_{i}^{2}.
\end{equation}
The default is $r=0$, giving a flat weighting after learning rate warmup. The use of the max of the squared learning rate seen so far is a heuristic choice. The weighting of the time step by power $r$ is theoretically justified, as a convergence rate can be proven for any $r \geq 0$, and for certain problem classes, improved rates are obtained for $r=1$ when using iterate averaging approaches \citep{defazio2021factorial}.

Figure~\ref{fig:anneal_r} shows a sweep of $r$ values for a 200B token training run, also using the $\beta$ annealing technique previously demonstrated in Figure~\ref{fig:sf_beta1-anneal}. All other hyper-parameters were kept constant. We see that $r=0$ (Equal weighting) is slightly better for runs shorter than 30B tokens, but after that $r=1$ has a very clear and significant advantage. Larger $r$ values than $r=1$ do not seem to be beneficial. Since $r=1$ values are typically used for non-strongly convex problems to accelerate convergence, this value is a sensible default for longer runs. Weighting by $r=1$ when averaging equally spaced checkpoints was found by \citet{li2025model} to improve the performance of the resulting model. \textbf{We recommend using $r=1$ weighting with Schedule-Free for long duration training runs.}

\section{Schedule-Free training is HIGHLY predictable}

It is well known that stochastic convex optimization problems have a $1/\sqrt{t}$ worst-case convergence rate. In the case of non-smooth $G$-Lipschitz optimization with SGD iterates $z_t$, the bound takes the form:
\begin{equation}
\mathbb{E}f(\bar{z}_{t})-f_{*}\leq\frac{\left\Vert z_{1}-x_{*}\right\Vert G}{\sqrt{t}} \label{eq:classical-bound}
\end{equation}
When using classical training approaches, this inverse-square-root decay of the loss during training is not seen in practice during real training runs. Normally, this theory-practice gap is hand-waved away, with the theoretical rate considered unrealistic due to one or all of the following factors:
\begin{description}
    \item[Non-Convexity] It seems ``unreasonable'' that we should expect this convex optimization bound to apply to giant neural network optimization problems. Function value convergence is not even guaranteed for non-convex problems under typical assumptions.
    \item[Average Iterate] This convergence rate is for the average iterate $\bar{z}_t$, not the actual iterates $z_t$. In practice for classical training approaches the average iterate is garbage, either it gives NaN, Inf, or terrible loss values. Classical analysis techniques give a $\log(t)$ worse rate for the last iterate \citep{shamir2013stochastic}, or use some sort of tail-averaging.
    \item[Worst-case behavior] This worst case bound is considered too pessimistic to hold in practice.
\end{description}
We believe that all of these points are misguided, and in fact when using Schedule-Free Learning, we should expect that our convergence behavior should follow Equation~\ref{eq:classical-bound} extremely closely, at least outside of an initial burn-in phase at the beginning of training.
\begin{figure}[t]
    \begin{subfigure}{0.5\textwidth}
        \includegraphics[width=\linewidth]{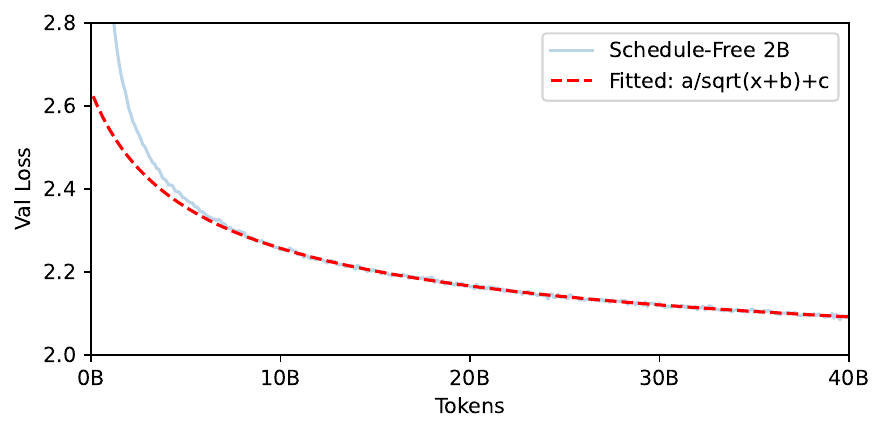}
        \caption{Schedule-Free Learning loss values fit an inverse-square-root decay rate very well outside an initial burn-in period}
        \label{fig:loss-fit}
    \end{subfigure}
    \hfill
    \begin{subfigure}{0.5\textwidth}
        \includegraphics[width=\textwidth]{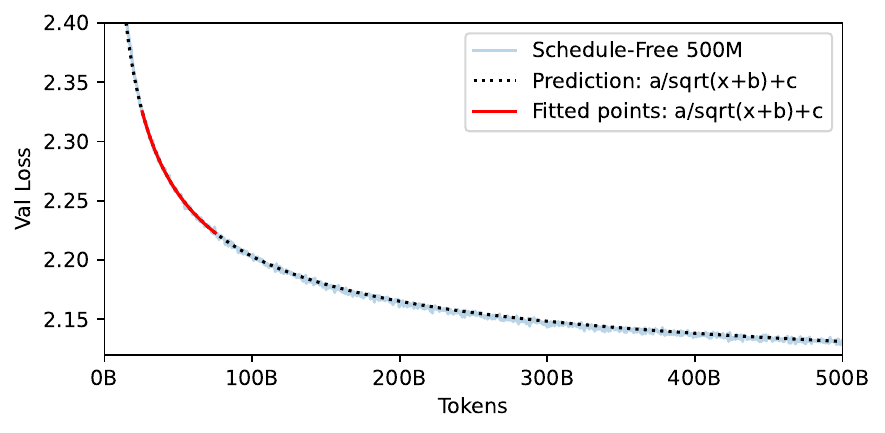}
        \caption{For long training runs, we can use early loss values to predict future loss values with remarkably high accuracy}
        \label{fig:loss-fit-long}
    \end{subfigure}
    \caption{\label{fig:curve-fit}Schedule-Free loss curves follow a predictable 1/sqrt decay}
\end{figure}
To verify this hypothesis, we performed a curve fitting procedure using \texttt{scipy.optimize.curve\_fit} with default parameters, fitting the following equation to the sequence of loss values within a single run:
\begin{equation}
f(x_{t})=\frac{a}{\sqrt{t+b}} + c
\end{equation}
Here $c$ approximates $f_*$, the optimal function value. The offset in the denominator accounts for a changing rate of convergence at the beginning due to learning rate warmup. In Figure~\ref{fig:loss-fit}, we apply this fit to the last $75\%$ of a short-ish duration (40B tokens) training run of a 2B model. Outside of the initial burn-in period, the fit is \textit{extremely} good. 

We can also use this kind of fitting to predict future loss values from early loss values during training. This is particularly useful for very long training runs. In Figure~\ref{fig:loss-fit-long}, we train a 500M parameter model for 500B tokens. We use the loss values from 5\% to 15\% run (dropping the initial 5\% burn-in period), to test if we can predict the entire loss curve from just 10\% of the training run. As we can see, this fit is also very good, \textbf{predicting near identical loss values to the actual measured loss values for the remainder of training.} The curve fit predicts that $f_*$ is $2.072$ for this $500M$ parameter model, compared to $1.887$ predicted for the 2B model.
\clearpage
\part*{Scaling Ladders}

We call the combination of the above approaches the ScheduleFree+ method. We ran a series of scaling ladder experiments, where at each scale we compare against a state-of-the-art Linear Decay schedule, using a learning rate tuned with a grid search. The Schedule-Free version has no learning rate grid search and relies purely on the Polyak Step-size at all scales. The reference implementation of ScheduleFree+ used for these experiments is available at \url{https://github.com/facebookresearch/schedule_free/blob/main/schedulefree/adamc_schedulefree_plus_paper.py}.

\subsection{Long Duration Training Runs}
\begin{figure}[b]
    \vspace{-1em}
    \centering
    \newcommand{\sideLabelWidth}{0.03} 
    \newcommand{\plotAreaWidth}{0.90}  
    \newcommand{\individualPlotWidth}{0.5} 
    \newcommand{\rowSpacing}{-0.5em}    

    \begin{tabular}{m{\sideLabelWidth\linewidth} l}
        \rotatebox[origin=c]{90}{\small (a) 120m} & 
        \begin{minipage}{\plotAreaWidth\linewidth}
            \includegraphics[width=\individualPlotWidth\linewidth]{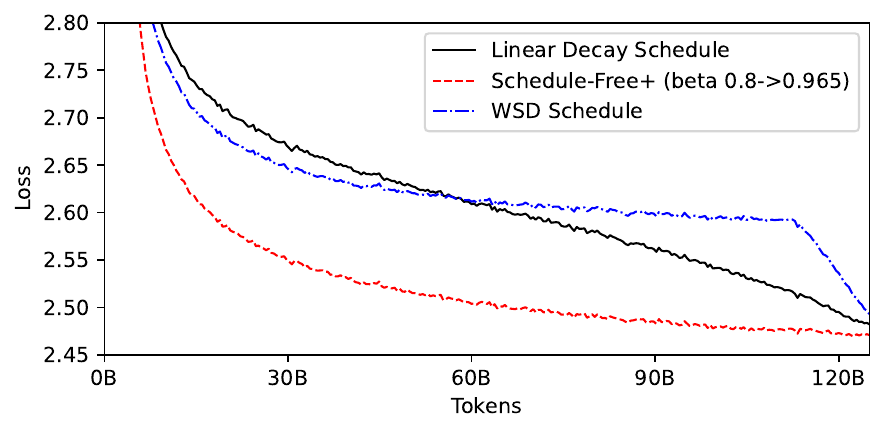}
            \includegraphics[width=\individualPlotWidth\linewidth]{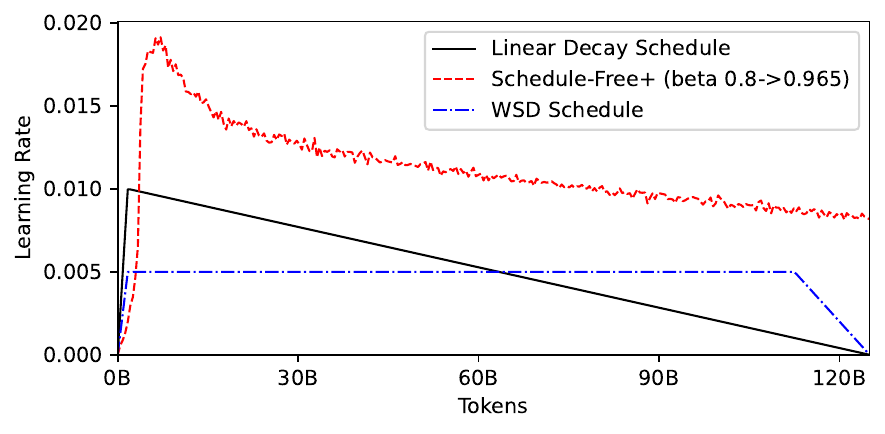}
        \end{minipage} \\ [\rowSpacing]

        \rotatebox[origin=c]{90}{\small (b) 250m} & 
        \begin{minipage}{\plotAreaWidth\linewidth}
            \includegraphics[width=\individualPlotWidth\linewidth]{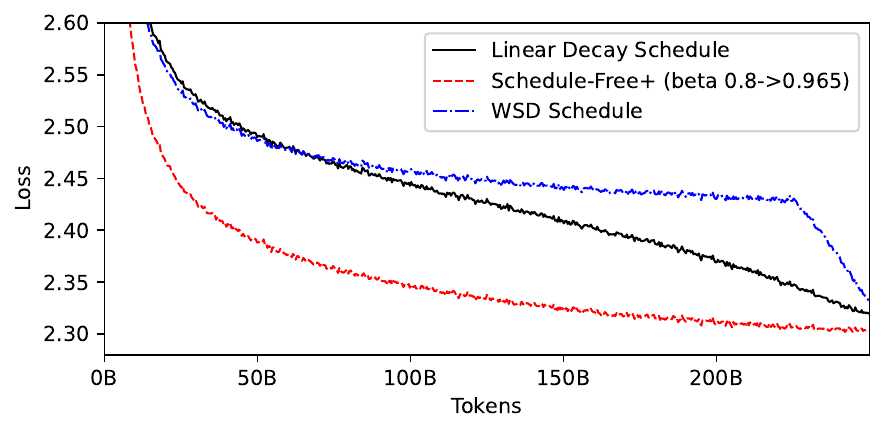}
            \includegraphics[width=\individualPlotWidth\linewidth]{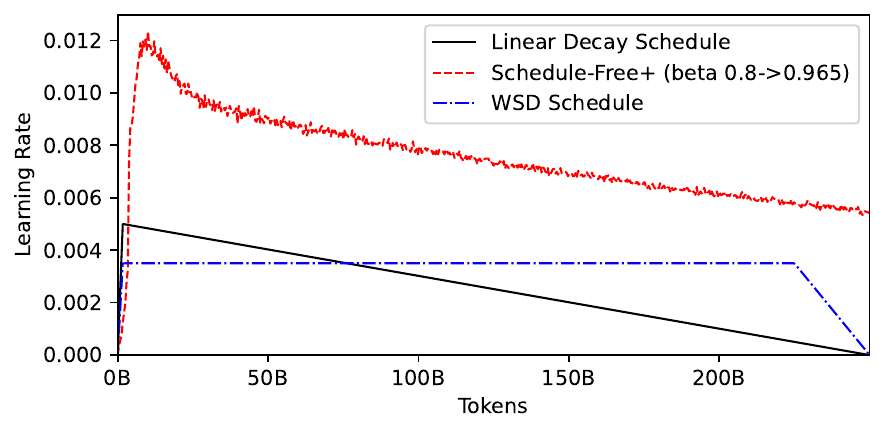}
        \end{minipage} \\ [\rowSpacing]

        \rotatebox[origin=c]{90}{\small (c) 500m} & 
        \begin{minipage}{\plotAreaWidth\linewidth}
            \includegraphics[width=\individualPlotWidth\linewidth]{figures/ladder/ladder_500m_1000x_anneal_loss.pdf}
            \includegraphics[width=\individualPlotWidth\linewidth]{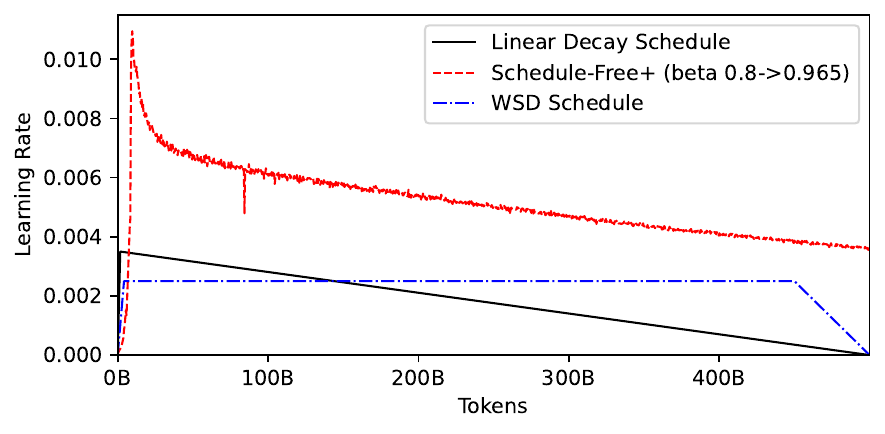}
        \end{minipage} \\ [\rowSpacing]

        \rotatebox[origin=c]{90}{\small (d) 1B} & 
        \begin{minipage}{\plotAreaWidth\linewidth}
            \includegraphics[width=\individualPlotWidth\linewidth]{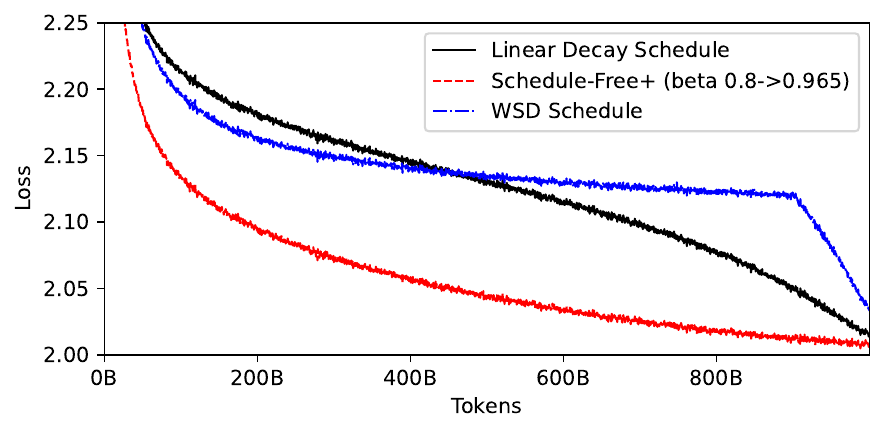}
            \includegraphics[width=\individualPlotWidth\linewidth]{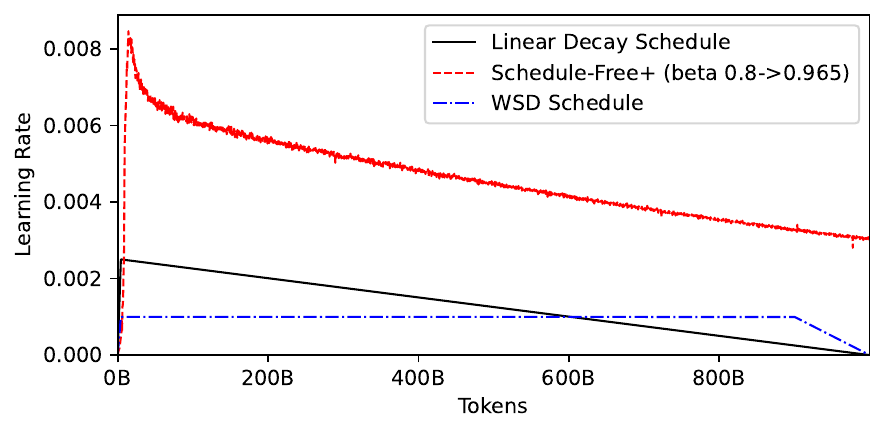}
        \end{minipage}
    \end{tabular}
    \vspace{-1em}
    \caption{1000 tokens per parameter training runs. Rows (a)-(d) show scaling from 120m to 1B parameters.}
    \label{fig:annealing-1000}
    \vspace{-1em}
\end{figure}

We found the greatest improvement over schedule-based approaches when training on large numbers of tokens, where beta-annealing and using r=1 weighted averaging (Section~\ref{sec:annealing-and-weighting}) becomes effective. Figure~\ref{fig:annealing-1000} shows the loss curves and adaptive step sizes at three increasing scales, using 1000 tokens per parameter. This tpp scale is significantly longer than the Chinchilla recommended 20 tokens per parameter, but is more representative of modern training regimes. Tokens per (active) parameter for frontier open models have increased at a rate of $>$3x per year \citep{epoch2025trainingtokensperparameter}.
ScheduleFree+ significantly improves upon the AdamW baseline both in terms of final loss, as well as intermediate loss values. Even without beta-annealing, ScheduleFree+ outperforms schedule-based approaches consistently for longer duration training runs (Figure~\ref{fig:noanneal-1000tpp}). For example, for the 120M model, we ran a grid of longer duration runs to see how long it would take to reach a comparable loss value using the linear decay schedule. \textbf{ScheduleFree+ reached the same final loss value as a 45\% longer run,} a 31\% reduction in training time to reach the same loss.

\begin{table}[h]
\centering
\begin{tabular}{|c|c|c|c|c|c|c|c|}
\hline 
Parameters & Tokens & Steps & Batch size & GPUs & Warmup & Beta & r\tabularnewline
\hline 
\hline 
120M & 125B & 29,803 & 4M & 8 & 400 & $0.8\rightarrow 0.965$ & 1\tabularnewline
\hline 
250M & 250B & 59,605 & 4M & 32 & 400 & $0.8\rightarrow 0.965$ & 1\tabularnewline
\hline 
500M & 500B & 119,210 & 4M & 128 & 1000 & $0.8\rightarrow 0.965$ & 1\tabularnewline
\hline 
1B & 1T & 238,419 & 4M & 256 & 1000 & $0.8\rightarrow 0.965$ & 1\tabularnewline
\hline 
\end{tabular}\par
\caption{1000 tokens per parameter settings}
\end{table}

\subsection{Shorter Duration Training Runs}
We further tested Schedule-Free at 100 tokens per parameter. At this scale beta-annealing is not effective, and we tested ScheduleFree+ with a fixed $\beta=0.9$. For models from 120M up to 2B parameters, we find that Schedule-Free exceeds the performance of the WSD schedule at all scales (Figure~\ref{fig:100tpp}). It also does well against the linear decay schedule, outperforming it at all but the 2B scale.

\begin{table}[h]
\centering
\begin{tabular}{|c|c|c|c|c|c|c|c|}
\hline 
Parameters & Tokens & Steps & Batch size & GPUs & Warmup & Beta & r\tabularnewline
\hline 
\hline 
120M & 12.5B & 2,981 & 4M & 8 & 400 & 0.9 & 0\tabularnewline
\hline 
250M & 25B & 5,961 & 4M & 8 & 400 & 0.9 & 0\tabularnewline
\hline 
500M & 50B & 11,921 & 4M & 8 & 1000 & 0.9 & 0\tabularnewline
\hline 
1B & 100B & 23,842 & 4M & 32 & 2000 & 0.9 & 0\tabularnewline
\hline 
2B & 200B & 47,684 & 4M & 128 & 2000 & 0.9 & 0\tabularnewline
\hline 
\end{tabular}\par
\caption{100 tokens per parameter settings}
\end{table}

Using 20 tokens per parameter still shows an advantage for ScheduleFree+ at smaller model sizes, but we see a small performance gap for larger models when compared to the linear decay schedule. At this token budget, the loss curves for Schedule-Free Learning are not significantly below those from a schedule, which may also contribute to the differences. ScheduleFree+ does however outperform the WSD schedule at all model scales. 

At this tpp scale the duration of the run for which gradient norms and weight norms are not at steady state is close to 50\% of the run. Schedule-Free takes longer to hit steady state and this contributes to the lower performance for these very short runs. Initializing the weights closer to steady state or using weight projection would help here.

\begin{table}[h]
\centering
\begin{tabular}{|c|c|c|c|c|c|c|c|}
\hline 
Parameters & Tokens & Steps & Batch size & GPUs & Warmup & Beta & r\tabularnewline
\hline 
\hline 
120M & 2.4B & 2,289 & 1M & 8 & 400 & 0.9 & 0\tabularnewline
\hline 
250M & 5B & 4,769 & 1M & 8 & 400 & 0.9 & 0\tabularnewline
\hline 
500M & 10B & 9,537 & 1M & 8 & 1000 & 0.9 & 0\tabularnewline
\hline 
1B & 20B & 19,074 & 1M & 8 & 2000 & 0.9 & 0\tabularnewline
\hline 
2B & 40B & 38,147 & 1M & 32 & 2000 & 0.9 & 0\tabularnewline
\hline 
\end{tabular}\par
\caption{20 tokens per parameter settings}
\end{table}

\afterpage{
\clearpage
\begin{figure}[p]
    \begin{subfigure}{\linewidth}
        \includegraphics[width=0.49\linewidth]{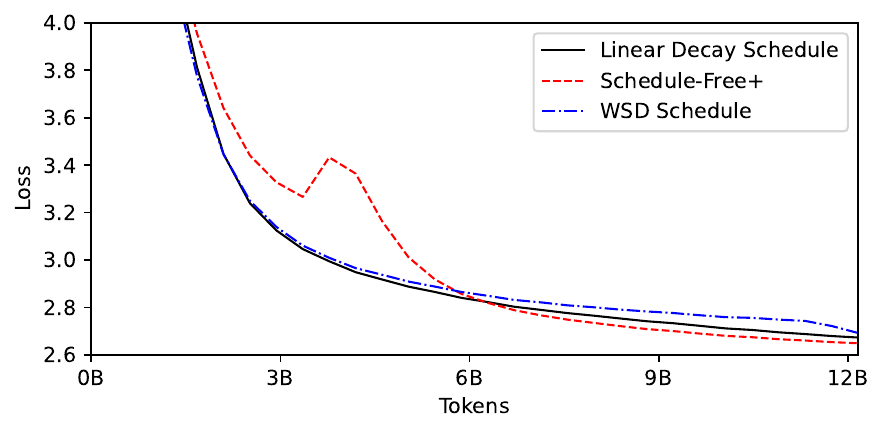}
        \includegraphics[width=0.49\linewidth]{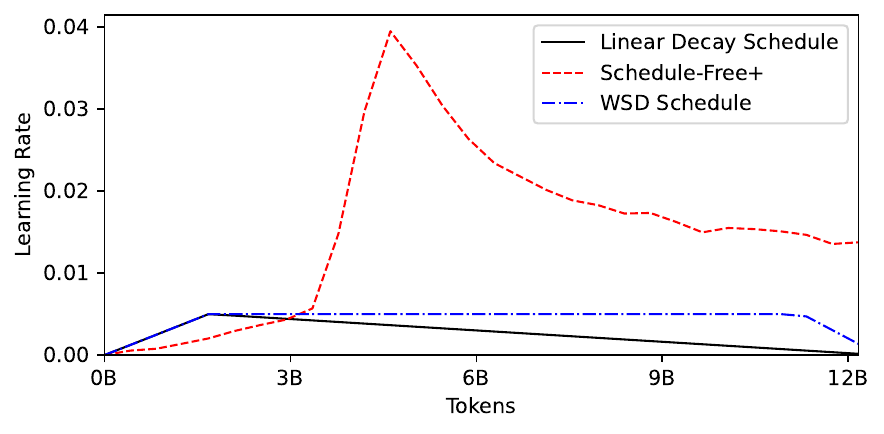}
        \vspace{-1em}\caption{120m parameters}
    \end{subfigure}
    \begin{subfigure}{\linewidth}
        \includegraphics[width=0.49\linewidth]{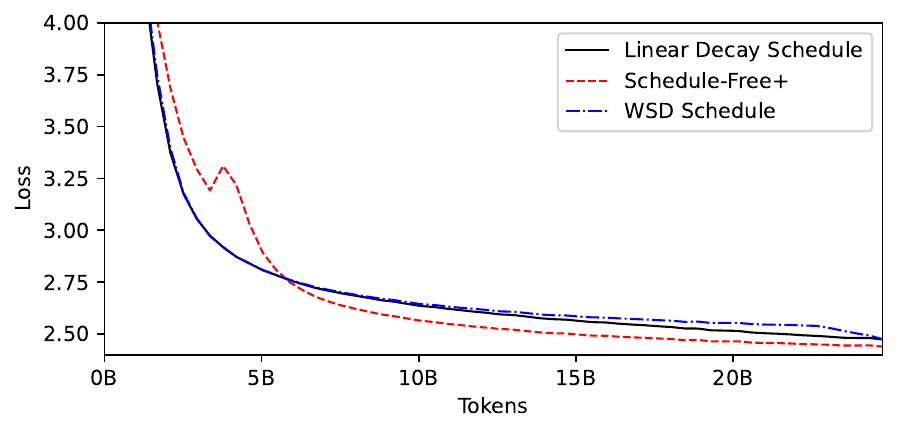}
        \includegraphics[width=0.49\linewidth]{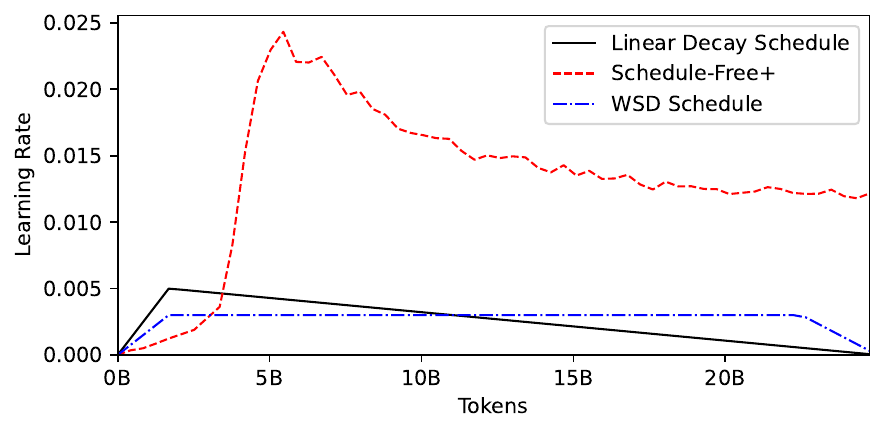}
        \vspace{-1em}\caption{250m parameters}
    \end{subfigure}
    \begin{subfigure}{\linewidth}
        \includegraphics[width=0.49\linewidth]{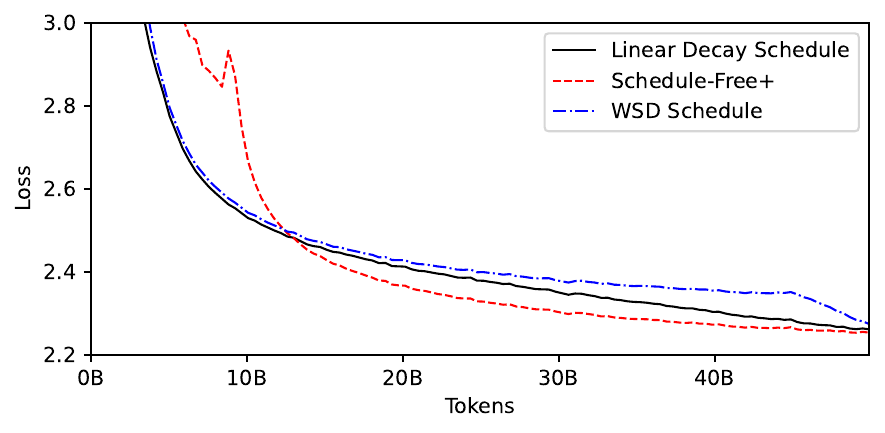}
        \includegraphics[width=0.49\linewidth]{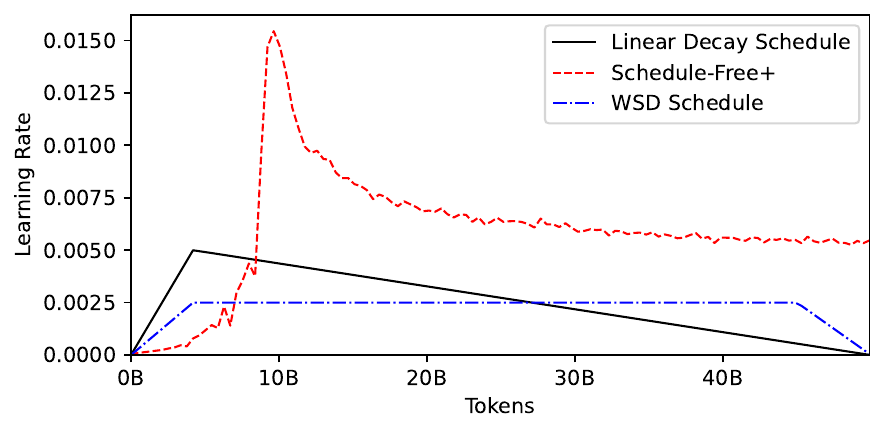}
        \vspace{-1em}\caption{500m parameters}
    \end{subfigure}
    \begin{subfigure}{\linewidth}
        \includegraphics[width=0.49\linewidth]{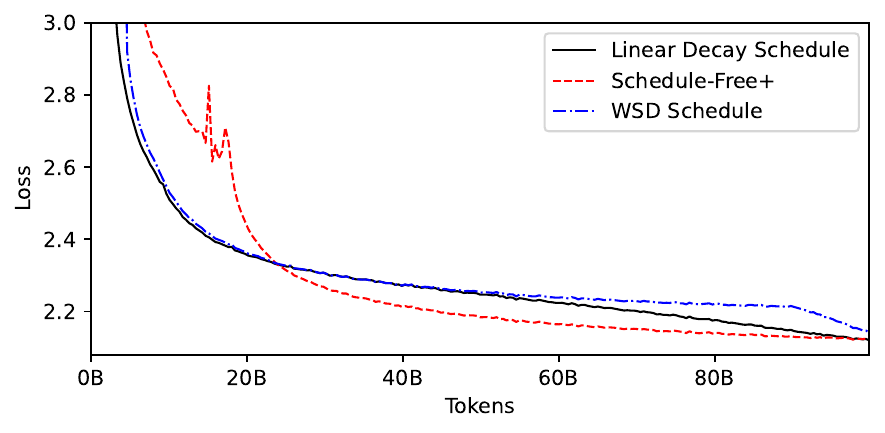}
        \includegraphics[width=0.49\linewidth]{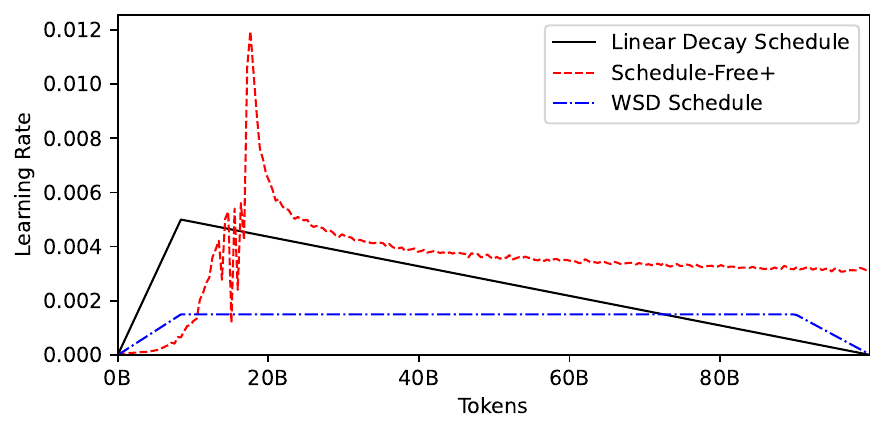}
        \vspace{-1em}\caption{1b parameters}
    \end{subfigure}
    \begin{subfigure}{\linewidth}
        \includegraphics[width=0.49\linewidth]{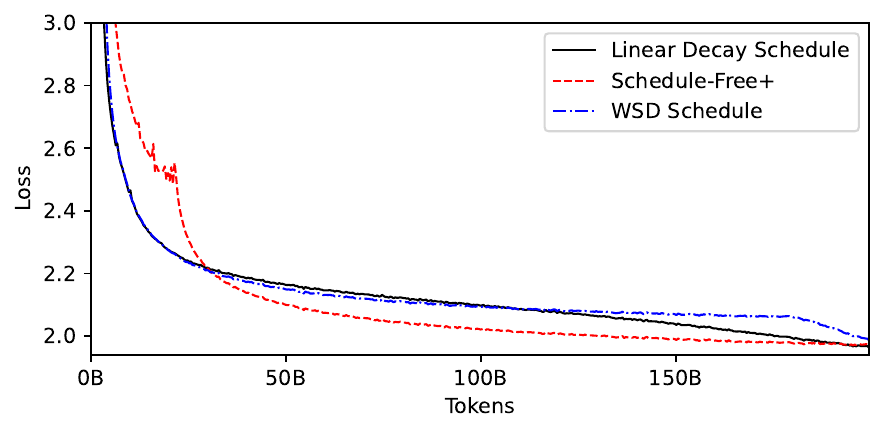}
        \includegraphics[width=0.49\linewidth]{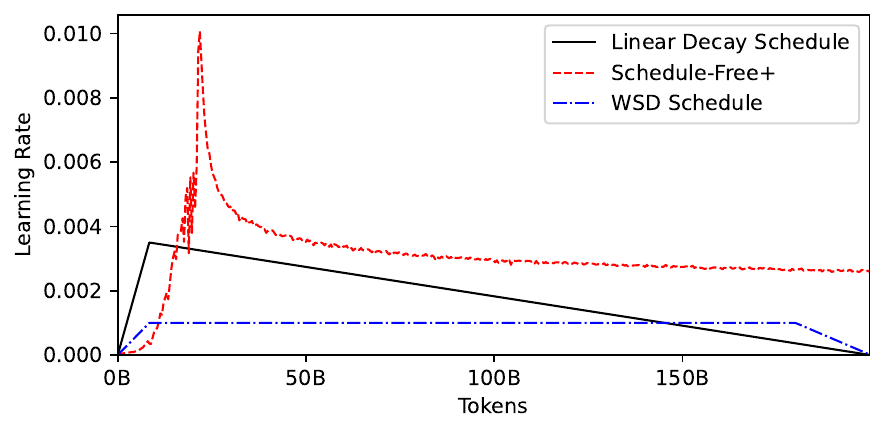}
        \vspace{-1em}\caption{2b parameters}
    \end{subfigure}
    \caption{\label{fig:100tpp}100 tokens per parameter}
\end{figure}
\clearpage
}
\afterpage{
\clearpage
\begin{figure}[p]
    \begin{subfigure}{\linewidth}
        \includegraphics[width=0.49\linewidth]{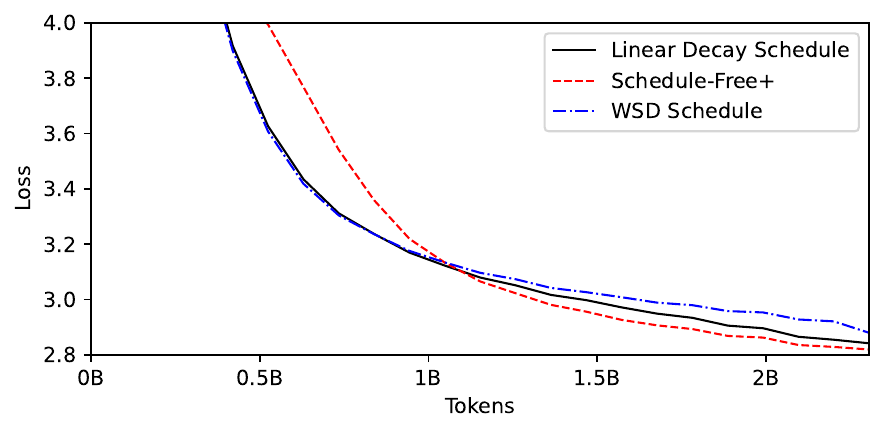}
        \includegraphics[width=0.49\linewidth]{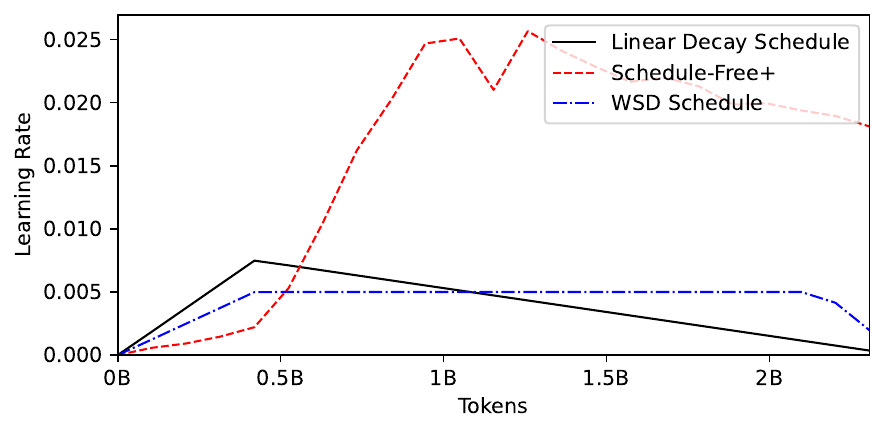}
        \vspace{-1em}\caption{120m parameters}
    \end{subfigure}
    \begin{subfigure}{\linewidth}
        \includegraphics[width=0.49\linewidth]{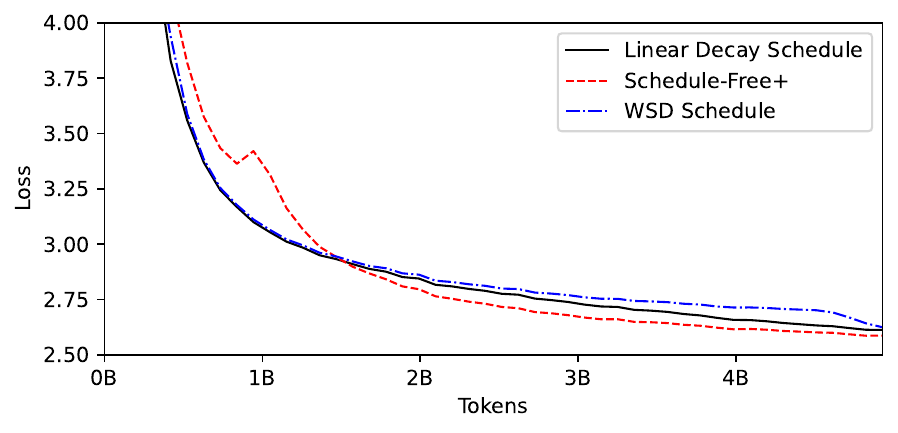}
        \includegraphics[width=0.49\linewidth]{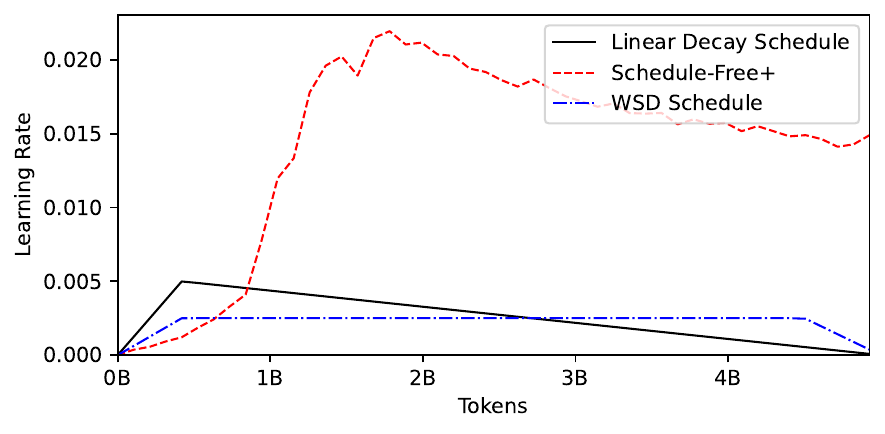}
        \vspace{-1em}\caption{250m parameters}
    \end{subfigure}
    \begin{subfigure}{\linewidth}
        \includegraphics[width=0.49\linewidth]{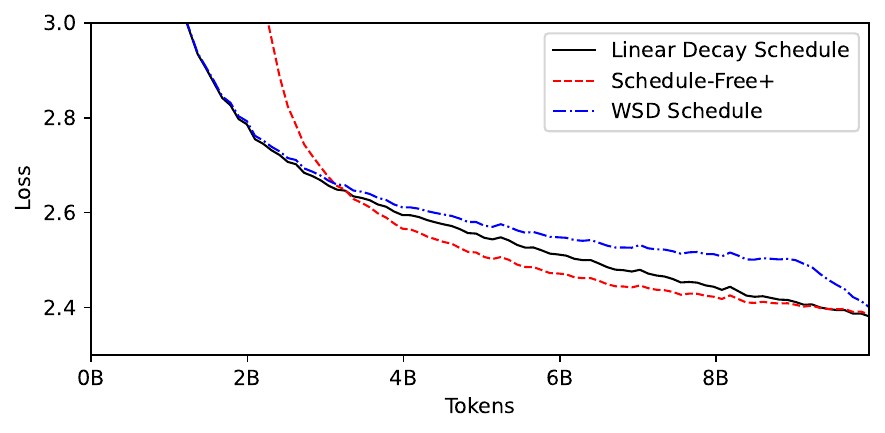}
        \includegraphics[width=0.49\linewidth]{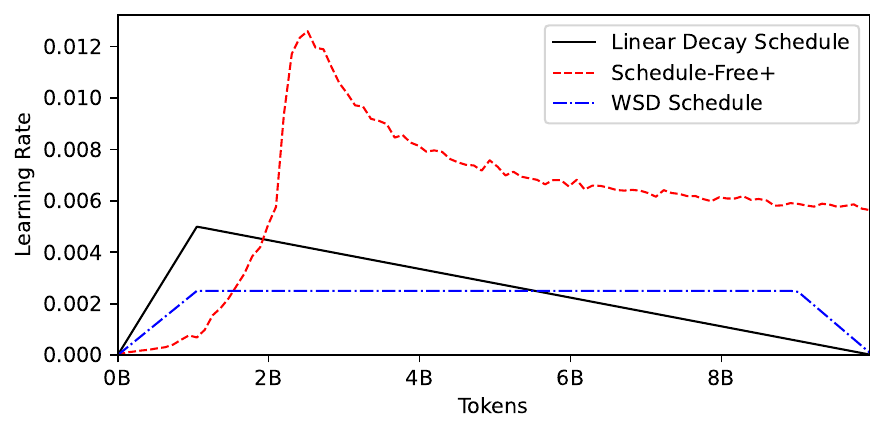}
        \vspace{-1em}\caption{500m parameters}
    \end{subfigure}
    \begin{subfigure}{\linewidth}
        \includegraphics[width=0.49\linewidth]{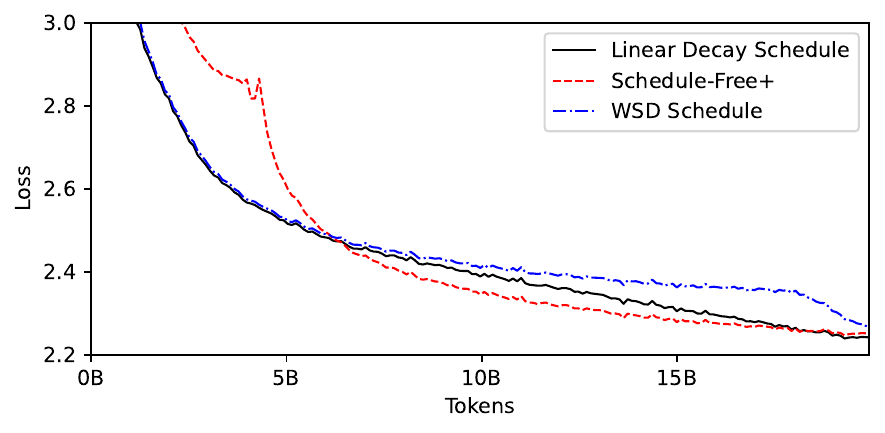}
        \includegraphics[width=0.49\linewidth]{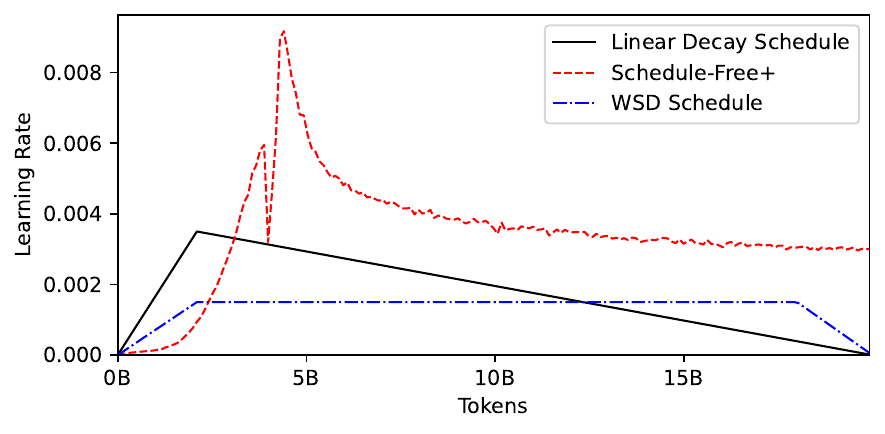}
        \vspace{-1em}\caption{1b parameters}
    \end{subfigure}
    \begin{subfigure}{\linewidth}
        \includegraphics[width=0.49\linewidth]{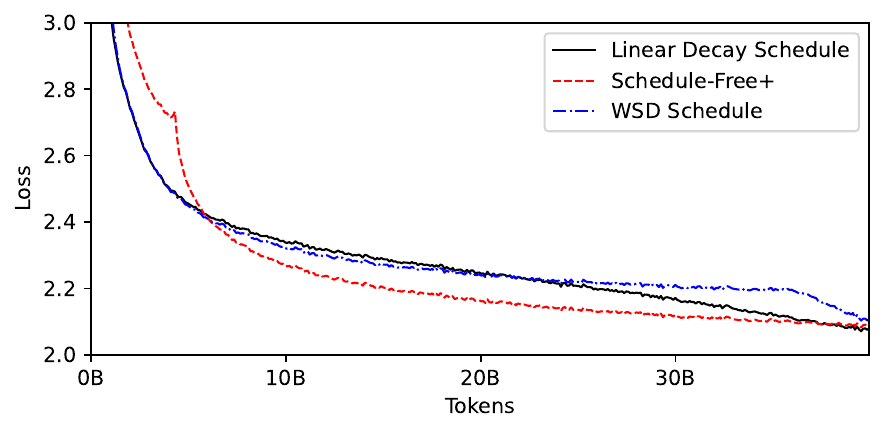}
        \includegraphics[width=0.49\linewidth]{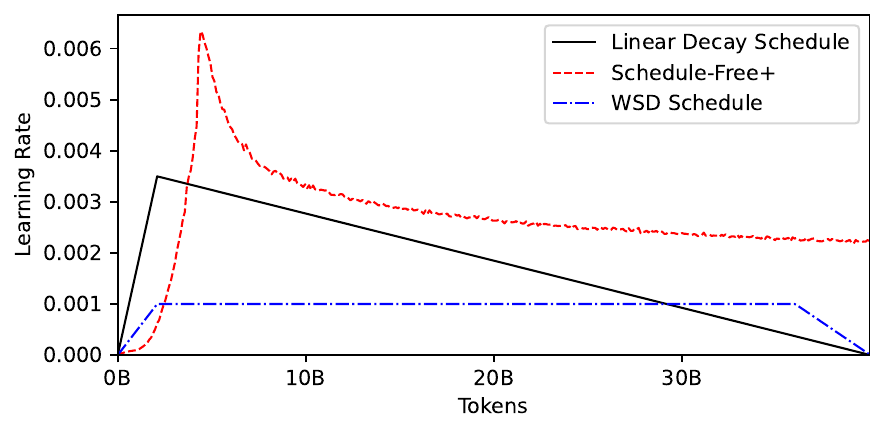}
        \vspace{-1em}\caption{2b parameters}
    \end{subfigure}
    \caption{20 tokens per parameter training runs}
\end{figure}
\clearpage
}
\clearpage
\begin{algorithm}[H]
\caption{ScheduleFree+ (AdamC + Schedule-Free + Polyak)}
\label{alg:schedulefree_plus}
\begin{algorithmic}[1]
\Require Warmup schedule $\gamma_t$, weight decay $\lambda$, Adam betas $(\beta_1, \beta_2)$, numerical stability constant $\epsilon$
\Require Schedule-Free parameters $r$, $p$ (\texttt{weight\_lr\_power}), $C_{\text{warmup}}$, $\beta_{\text{sf}}$, $\beta_{\text{sf}}^{\max}$, $T_{\text{anneal}}$
\Require Polyak EMA coefficient $\beta_{p}$
\State Initialize parameters $x_0 = y_0 = z_0 = \theta_0$
\State Initialize Adam moments $m_0 = 0$, $v_0 = 0$
\State Initialize Polyak EMA $e_0 = 0$, schedule weight sum $W_0 = 0$, max learning rate $\gamma_{\max} = \epsilon$
\For{$t = 1 \dots T$}
    \State \Comment{\textbf{1. Schedule-Free Momentum Annealing}}
    \If{$T_{\text{anneal}} > 0$}
        \State $\tau_t \gets \min(t / T_{\text{anneal}}, 1)$
        \State $\tilde{\beta}_t \gets 1 - \exp\left((1 - \tau_t) \ln(1 - \beta_{\text{sf}}) + \tau_t \ln(1 - \beta_{\text{sf}}^{\max})\right)$
    \Else
        \State $\tilde{\beta}_t \gets \beta_{\text{sf}}$
    \EndIf

    \State \Comment{\textbf{2. Global Metrics \& Polyak Step Size}}
    \State Evaluate gradient at $y_{t-1}$: $g_t \gets \nabla f(y_{t-1})$
    \State Get global function value: $F_t \gets f(y_{t-1})$
    \State Compute global gradient $L_1$ norm: $L_1 \gets \|g_t\|_1$
    \State Compute inner product correction: $I_t \gets \tilde{\beta}_t \langle g_t, z_{t-1} - x_{t-1} \rangle$
    
    \State Update Polyak gradient $L_1$ EMA: $e_t \gets \beta_p e_{t-1} + (1 - \beta_p) L_1 \sqrt{\pi / 2}$
    \State Bias-correct $L_1$ EMA: $\hat{e}_t \gets e_t / (1 - \beta_p^t)$
    \State Compute Polyak scalar: $\eta_t \gets \max(0, F_t + I_t) / \hat{e}_t$
    \State Compute effective learning rate: $\alpha_t \gets \gamma_t \cdot \eta_t$
    \State $\gamma_{\max} \gets \max(\alpha_t, \gamma_{\max})$

    \State \Comment{\textbf{3. Schedule-Free Averaging Weights}}
    \If{$t \le C_{\text{warmup}}$}
        \State $c_t \gets 1$
    \Else
        \State $w_t \gets t^r \cdot \gamma_{\max}^p$
        \State $W_t \gets W_{t-1} + w_t$
        \State $c_t \gets w_t / W_t$
    \EndIf

    \State \Comment{\textbf{4. Per-Parameter Updates}}
    \State $z_t \gets z_{t-1} - \alpha_t^2 \lambda y_{t-1}$ \hfill \Comment{Decoupled AdamC weight decay}
    \State $m_t \gets \beta_1 m_{t-1} + (1 - \beta_1) g_t$ \hfill \Comment{Update first moment}
    \State $v_t \gets \beta_2 v_{t-1} + (1 - \beta_2) g_t^2$ \hfill \Comment{Update second moment}
    \State $\hat{m}_t \gets m_t / (1 - \beta_1^t)$ \hfill \Comment{Bias correction}
    \State $\hat{v}_t \gets v_t / (1 - \beta_2^t)$
    
    \State $z_t \gets z_t - \alpha_t \frac{\hat{m}_t}{\sqrt{\hat{v}_t} + \epsilon}$ \hfill \Comment{Adam step on $z$}
    \State $x_t \gets (1 - c_t) x_{t-1} + c_t z_t$ \hfill \Comment{Update evaluation point $x$}
    \State $y_t \gets \tilde{\beta}_t x_t + (1 - \tilde{\beta}_t) z_t$ \hfill \Comment{Compute next query point $y$}
\EndFor
\State \Return $x_T$
\end{algorithmic}
\end{algorithm}
\clearpage
\part*{Discussion}
\section{Warmup Stable Decay Schedules}
The flexibility to halt training early, or to continue to train indefinitely, motivated the introduction of the WSD schedule \citep{Zhai_2022_CVPR, hu2024minicpm}. WSD schedules start with a standard learning-rate warmup (as used by both Schedule-Free and other schedules), followed by a constant learning rate thereafter. Training can be early stopped by introducing a linear learning rate anneal to 0, typically 10\% of the training run is allocated to this annealing phase. WSD-like schedules have been used in some frontier model training runs \citep{bi2024deepseek, kimiteam2025kimik2openagentic}.

Schedule-Free Learning gives true anytime training; it doesn't require any additional annealing, a significant advantage over WSD schedules. There is no delay while you wait for the annealing phase before having a usable model to run evaluations on. Restarting training is also much more straight-forward without annealing.

The efficacy of WSD schedules has recently been motivated by a river-valley view of the loss landscape \citep{wen2025understanding}. In this view, even when the loss curve is not decreasing progress is still being made along `rivers' in the loss landscape, and this progress is revealed during the annealing phase as the high loss due to bouncing off of valley-walls is reduced. Schedule-Free Learning has also recently been considered from this viewpoint \citep{song2025through}, where it's argued that it navigates rivers without the high-loss bouncing that occurs with the WSD schedule.

It is tempting to believe that the river-valley landscape implies a twisting, winding river and rugged ever-changing mountain ranges, a complex landscape out of reach theoretical analysis. Remarkably, it appears that convex optimization theory can precisely explain the shape of loss curves that arise from the WSD schedule. \citet{schaipp2025the} show that the precise shape of the loss curve can be predicted by the use of schedule-dependent learning rate bounds. Given a schedule $\eta_t$ with peak learning rate $\gamma$, \citet{defazio2023when} show that the loss at every time-step is bounded by the following (rather complicated) expression: 
\begin{equation}
    E\left[f(x_{t})-f_{*}\right]\leq\frac{1}{2\gamma\bar{\eta}_{t}}\left[D^{2}+\gamma^{2}\sum_{i=1}^{t}\eta_{i}^{2}E\left\Vert g_{i}\right\Vert ^{2}\right]+\frac{\gamma}{2}\sum_{k=1}^{t-1}\frac{\eta_{k}}{\sum_{i=k+1}^{t}\eta_{i}}\left(\frac{1}{\sum_{i=k}^{t}\eta_{i}}\sum_{i=k}^{t}\eta_{i}^{2}E\left\Vert g_{i}\right\Vert ^{2}\right), \label{eq:anytime-loss}
\end{equation}
where $\bar{\eta}_{t}=\sum_{i=1}^{t}\eta_{i}$ and $D$ is an estimate of the initial distance to the solution. The right hand side depends on the expected gradient norms at each time-step, however assuming constant gradient norms gives good predictions outside of the initial warmup-phase of optimization. When evaluated numerically, this bound predicts the same loss drops during the annealing phase of the WSD that are seen in practice.

We use the Linear Decay schedule as the baseline horizon-dependent experiments in this work. The Linear Decay schedule arises from optimizing the right-hand-side of Equation~\ref{eq:anytime-loss} to find the optimal schedule when expected gradient norms are flat during training, and \citet{defazio2023when} show that it is a worst case-optimal schedule for non-smooth convex optimization problems. Besides this theoretical motivation, Linear Decay schedules have been observed to give state-of-the-art results across a wide variety of problems, including LLMs \citep{bergsma2025straight} with just a few exceptions such as long duration training runs on ResNet models where classical cosine schedules works well. In the convex case, the river-valley view suggests a small, long (but convex) river surrounded by mountains of varying steepness. The loss landscape along the WSD basin has been empirically observed to exhibit (quasi-)convexity \citep{belloni2025universal}

The WSD schedule's practical performance is good, but it is consistently outperformed by the linear decay schedule in our experiments when tuned to the same training horizon. The suboptimality of the WSD schedule against properly tuned baseline schedules is not emphasized in the literature, but has been observed by others also \citep{Zhai_2022_CVPR, bergsma2025straight}. 

\section{Cosine Schedules}
\begin{figure}[t]
    \begin{subfigure}{0.5\textwidth}
        \includegraphics[width=\linewidth]{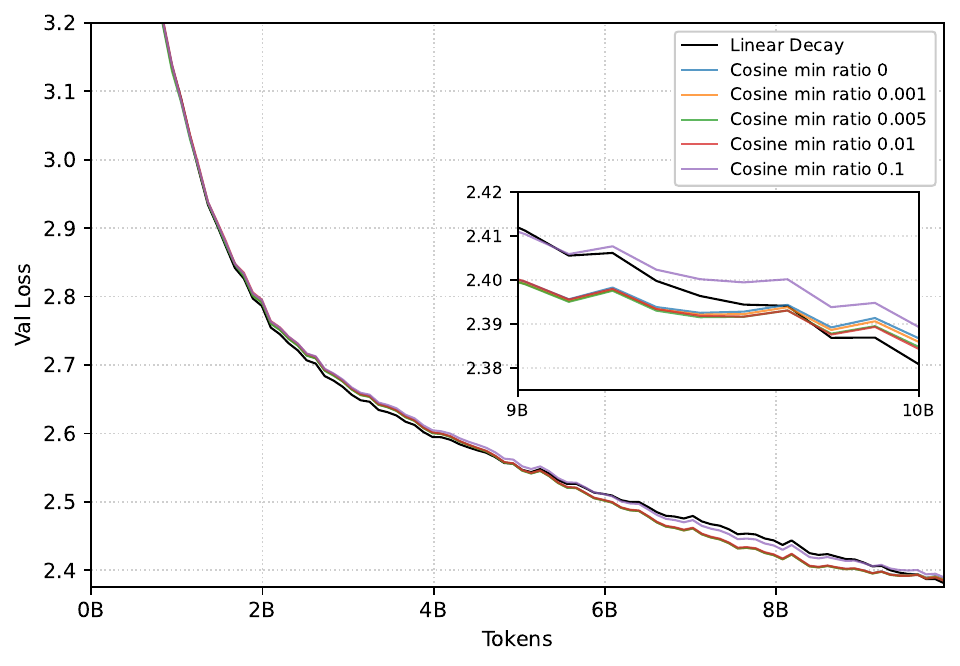}
        \caption{500M model at 20 tokens per parameter}
    \end{subfigure}
    \hfill
    \begin{subfigure}{0.5\textwidth}
        \includegraphics[width=\textwidth]{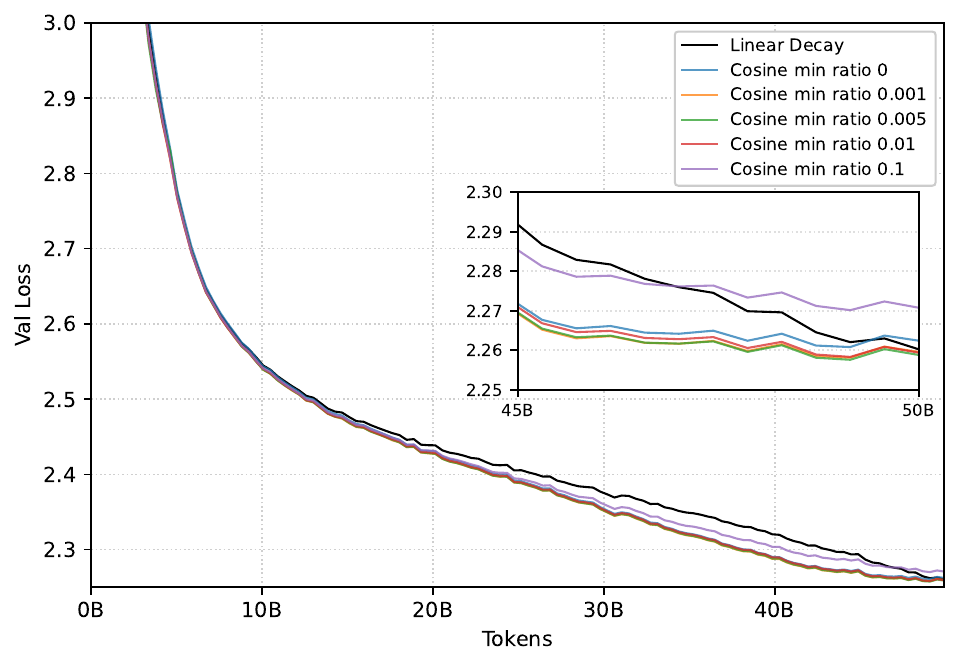}
        \caption{500M model at 100 tokens per parameter}
    \end{subfigure}
    \caption{\label{fig:cosine}Cosine schedules tie or even under-perform Linear Decay schedules for LLM training}
\end{figure}

In the early era of LLM training, a cosine schedule that decreases the learning rate from its initial optimal value to a 10x smaller value was heavily used,  due to its appearance in a series of influential DeepMind papers \citep{rae2021scaling, hoffmann2022empirical}. Figure~\ref{fig:cosine} shows a comparison of Linear Decay Schedules and Cosine Schedules when both are tuned by learning rate grid search. For the cosine schedule, we see that a 0.1 ratio between max and min is suboptimal, and there are only tiny differences between the other ratios tested, only visible in the magnified loss plots. 

\section{Model Merging, Souping and Averaging}
Instead of decreasing the learning rate during the later stages of training, multiple checkpoints can be averaged to obtain a lower loss point. This approach was first proposed (to our knowledge) by \citet{sandler2023training}, and was more recently explored at small scale by \citet{NEURIPS2024_8b970e15} and at larger scales by \citet{li2025model}. Averaging of model checkpoints has recently been popularized under the name \emph{Model Merging}, but the general idea is closely related to stochastic weight averaging \citep{Izmailov2018AveragingWL} and classical Polyak averaging \citep{polyak1990new, ruppert} explored in the convex optimization literature. This link to optimization theory is increasingly de-emphasized in more recent literature, or even lost, yet it provides a crucial theoretical underpinning for these methods.

Weight averaging of neural networks is problematic given the highly nonconvex nature of the loss landscape, and obviously shouldn't work. Having said that, it has been repeatedly demonstrated to work well in practice \citep{yang2024model}. Averages can even compete with ensembles of multiple models, as demonstrated in the Model Souping approach \citep{pmlr-v162-wortsman22a}. Averaging of base models with their fine-tuned counterparts during post-training has been shown to improve generalization performance \citep{Wortsman_2022_CVPR, NEURIPS2022_bc6cddcd}. Averaging can used to increase parallelism during the training process, with models being trained independently for 100s of steps before merging, as used by DiLoCo  \citep{douillard2023diloco}, or for longer durations on domain-specific data, as done in the Branch-Train-Merge approach \citep{li2022branch}.

Schedule-Free differs from the above approaches by \emph{mixing-in} the average point during training; Polyak-like averaging procedures use the average point in a purely off-line fashion, a crucial difference. Without this mixing operation, Schedule-Free Learning performs extremely poorly; it is an absolutely critical component of the method. Checkpoint averaging approaches typically focus on averaging points from the later parts of the training run (sometimes known as tail or suffix averaging), which is more stable. In contrast, Schedule-Free Learning's interpolation appears to enable averaging of earlier iterates, and generally yields faster convergence.

\section*{Conclusion}
We demonstrate that Schedule-Free Learning can be adapted to perform well on Large Language Model training tasks, providing a stable, anytime training procedure that outperforms both other anytime procedures such as WSD schedule-training, and also state-of-the-art horizon aware training approaches. The modifications here address limitations of Schedule-Free Learning identified by \citet{NEURIPS2024_8b970e15} and \citet{morwani2025connections}. Furthermore, in combination with Adaptive step sizes given by a novel modification of the Polyak step size, we demonstrate fully learning-rate-free learning scales to LLMs.

\clearpage
\bibliography{main}
\bibliographystyle{apalike}
\newpage
\appendix

\section{Additional Plots}

\begin{figure}
    \begin{subfigure}{\linewidth}
        \includegraphics[width=0.49\linewidth]{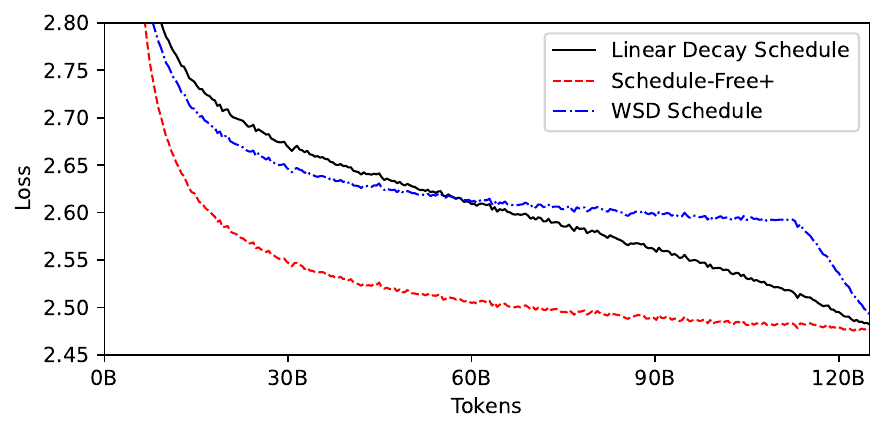}
        \includegraphics[width=0.49\linewidth]{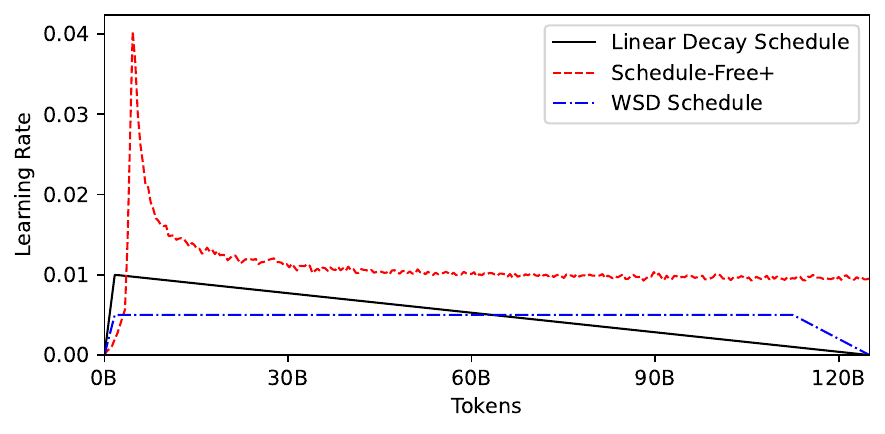}
        \vspace{-1em}\caption{120m parameters}
    \end{subfigure}
    \begin{subfigure}{\linewidth}
        \includegraphics[width=0.49\linewidth]{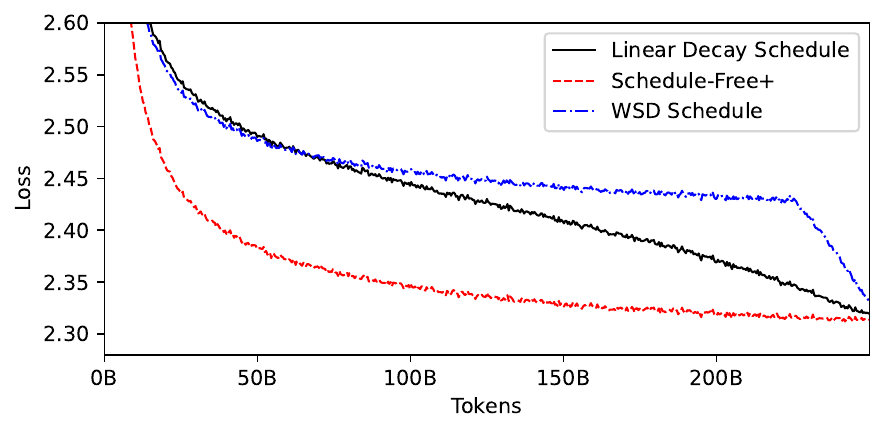}
        \includegraphics[width=0.49\linewidth]{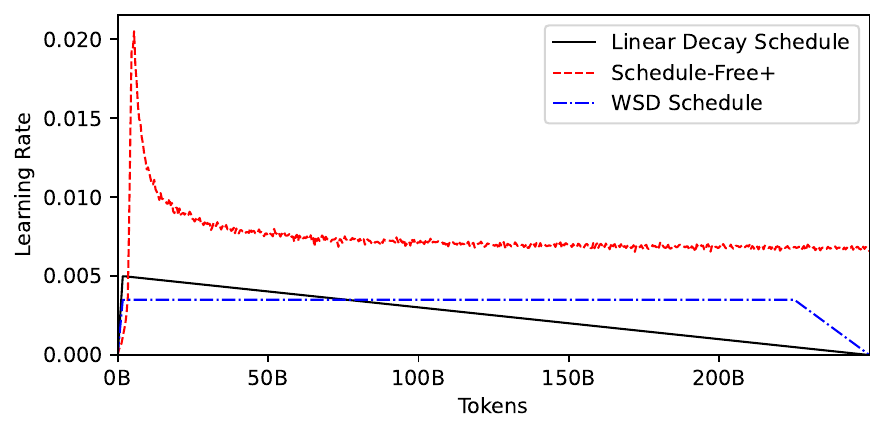}
        \vspace{-1em}\caption{250m parameters}
    \end{subfigure}
    \begin{subfigure}{\linewidth}
        \includegraphics[width=0.49\linewidth]{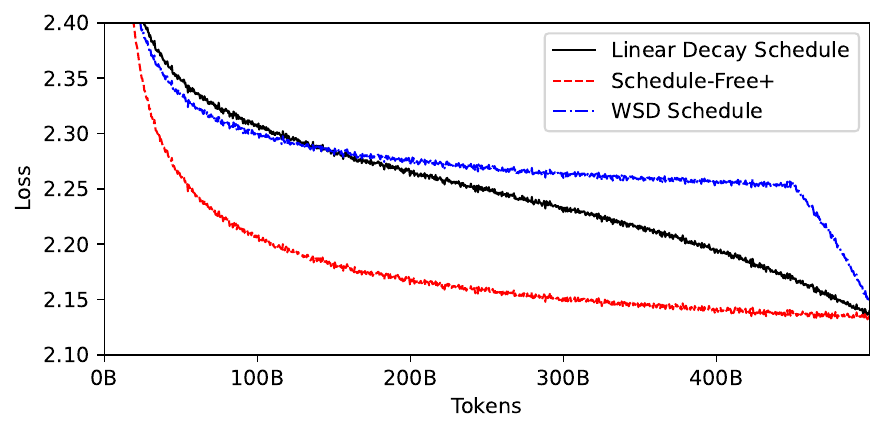}
        \includegraphics[width=0.49\linewidth]{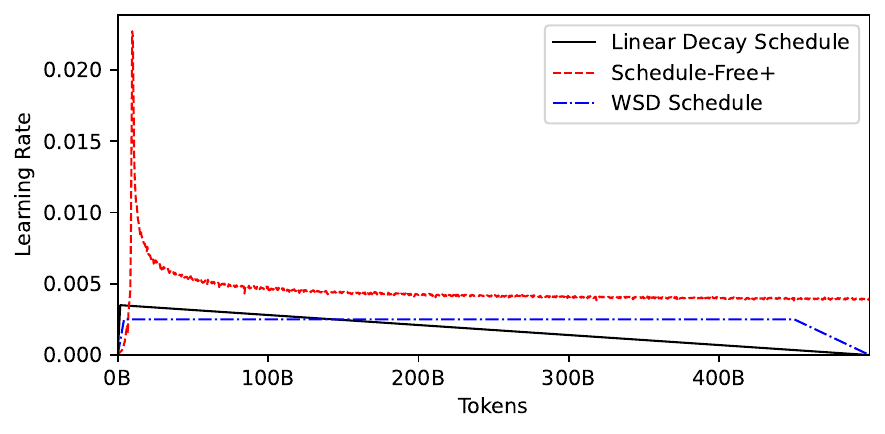}
        \vspace{-1em}\caption{500m parameters}
    \end{subfigure}
    \caption{\label{fig:noanneal-1000tpp}1000 tokens per parameter without beta-annealing}
\end{figure}

\section{Model Architecture Settings}
\begin{tabular}{|c|c|c|c|c|c|}
\hline 
 & 120M & 250M & 500M & 1B & 2B\tabularnewline
\hline 
\hline 
dim & 768 & 1024 & 1280 & 2048 & 2560\tabularnewline
\hline 
layers & 12 & 14 & 21 & 16 & 23\tabularnewline
\hline 
heads & 12 & 16 & 20 & 32 & 32\tabularnewline
\hline 
ffn exp & 4 & 4 & 4 & 4 & 4\tabularnewline
\hline 
ff multiplier & 256 & 256 & 256 & 256 & 256\tabularnewline
\hline 
rope theta & 10000 & 10000 & 10000 & 10000 & 10000\tabularnewline
\hline 
\end{tabular}

\section{Configuration Used for Each Figure}

\subsection{Figure~\ref{fig:intro}}
ScheduleFree+ runs here include the Polyak step size, warm starting, $r=1$ and beta annealing. A 500M parameter model was used. 

\begin{tabular}{|c|c|c|c|}
\hline 
Method & Linear Decay & WSD & ScheduleFree+\tabularnewline
\hline 
\hline 
Tokens & 500B & 500B & 500B\tabularnewline
\hline 
\hline 
Sequence Length & 1024 & 1024 & 1024 \tabularnewline
\hline 
Batch-size & 4M & 4M & 4M\tabularnewline
\hline 
GPUs & 128 & 128 & 128\tabularnewline
\hline 
Steps & 119,210 & 119,210 & 119,210\tabularnewline
\hline 
Warmup & 1000 & 1000 & 1000\tabularnewline
\hline 
Learning Rate & 0.0035 & 0.0025 & N/A\tabularnewline
\hline 
Decay & 0.05 & 0.05 & 20\tabularnewline
\hline 
Clipping & 1.0 & 1.0 & 1.0\tabularnewline
\hline 
r & N/A & N/A & 1.0\tabularnewline
\hline 
\end{tabular}

\subsection{Figure~\ref{fig:batchsize} a}
Batch-size was varied directly up to 128 samples per gpu, above that gradient acumulation was used to gather larger effective batch sizes. A 120M parameter model was used.

\begin{tabular}{|c|c|c|}
\hline 
Method & AdamW & Schedule-Free\tabularnewline
\hline 
\hline 
Tokens & 7.8B & 7.8B\tabularnewline
\hline 
Batch-size & Varies & Varies\tabularnewline
\hline 
Sequence Length & 1024 & 1024\tabularnewline
\hline 
GPUs & 8 & 8\tabularnewline
\hline 
Steps & 15,000 & 15,000\tabularnewline
\hline 
Warmup & 50,000/bs & 50,000/bs\tabularnewline
\hline 
Warm-started & Yes & Yes\tabularnewline
\hline 
Learning Rate & 0.01{*}sqrt(bs/64) & 0.02{*}sqrt(bs/32)\tabularnewline
\hline 
Decay & 0.05 & 0.002\tabularnewline
\hline 
Clipping & 10 & 10\tabularnewline
\hline 
Schedule & Linear Decay & N/A\tabularnewline
\hline 
SF Beta & N/A & 0.9\tabularnewline
\hline 
\end{tabular}

\subsection{Figure~\ref{fig:batchsize} b}

\begin{tabular}{|c|c|c|}
\hline 
Method & AdamW & Schedule-Free\tabularnewline
\hline 
\hline 
Tokens & 7.8B & 7.8B\tabularnewline
\hline 
Batch-size & Varies & Varies\tabularnewline
\hline 
Sequence Length & 1024 & 1024\tabularnewline
\hline 
GPUs & 8 & 8\tabularnewline
\hline 
Steps & 15,000 & 15,000\tabularnewline
\hline 
Warmup & 50,000/bs & 50,000/bs\tabularnewline
\hline 
Warm-started & Yes & Yes\tabularnewline
\hline 
Learning Rate & 0.01{*}sqrt(bs/64) & 0.015{*}sqrt(bs/32)\tabularnewline
\hline 
Decay & 0.05 & 0.002\tabularnewline
\hline 
Clipping & 10 & 10\tabularnewline
\hline 
Schedule & Linear Decay & N/A\tabularnewline
\hline 
SF Beta & N/A & 0.9\tabularnewline
\hline 
AdamW Beta1 & 0.9 & 0.75\tabularnewline
\hline 
\end{tabular}

\subsection{Figure~\ref{fig:inner-mom} a}
Batch size of 0.5M was used, with weight decay 0.005. Otherwise hyper-parameters match the previous figure.

\subsection{Figure~\ref{fig:inner-mom} b}

\begin{tabular}{|c|c|c|}
\hline 
Method & SF Refinement & SF Outer Mom\tabularnewline
\hline 
\hline 
Tokens & 7.8B & 7.8B\tabularnewline
\hline 
Batch-size & Varies & Varies\tabularnewline
\hline 
Sequence Length & 1024 & 1024\tabularnewline
\hline 
GPUs & 8 & 8\tabularnewline
\hline 
Steps & 15,000 & 15,000\tabularnewline
\hline 
Warmup & 50,000/bs & 50,000/bs\tabularnewline
\hline 
Warm-started & Yes & Yes\tabularnewline
\hline 
Learning Rate & 0.01{*}sqrt(bs/32) & 0.015{*}sqrt(bs/32)\tabularnewline
\hline 
Decay & 0.002 & 0.002\tabularnewline
\hline 
Clipping & 10 & 10\tabularnewline
\hline 
SF Beta & N/A & 0.9\tabularnewline
\hline 
AdamW Beta1 & 0.0 & 0.9\tabularnewline
\hline 
Refinement C & 50 & N/A\tabularnewline
\hline 
\end{tabular}

\subsection{Figure~\ref{fig:wd} a \& b}

A 120M parameter model was used.

\begin{tabular}{|c|c|c|c|c|}
\hline 
Method & AdamW & AdamC & SF WD 0.002 & SF WD 0.05 \tabularnewline
\hline 
\hline 
Tokens & 7.8B & 7.8B & 7.8B & 7.8B\tabularnewline
\hline 
Batch-size & 1M & 1M & 1M & 1M\tabularnewline
\hline 
Sequence Length & 1024 & 1024 & 1024 & 1024\tabularnewline
\hline 
GPUs & 8 & 8 & 8 & 8\tabularnewline
\hline 
Steps & 15,000 & 15,000 & 15,000 & 15,000\tabularnewline
\hline 
Warmup & 1000 & 1000 & 1000 & 1000\tabularnewline
\hline 
Warm-started & Yes & Yes & Yes & Yes\tabularnewline
\hline 
Learning Rate & 0.015 & 0.015 & 0.03 & 0.03\tabularnewline
\hline 
Decay & 0.05 & 0.05 & 0.002 & 0.05\tabularnewline
\hline 
Clipping & 10 & 10 & 10 & 10\tabularnewline
\hline 
Schedule & Linear Decay & Linear Decay & N/A & N/A\tabularnewline
\hline 
AdamW Beta & 0.9 & 0.9 & 0.75 & 0.75\tabularnewline
\hline 
SF Beta & N/A & N/A & 0.9 & 0.9\tabularnewline
\hline 
\end{tabular}

\subsection{Figure~\ref{fig:wd-weights} a, c \& d}
Weights norms are shown from Figure 4's runs.

\subsection{Figure~\ref{fig:l1}}
Configuration matches previous figure. 1/L1 runs use learning rate multiplier of 2.

\subsection{Figure~\ref{fig:l1_approximation}}

\begin{tabular}{|c|c|}
\hline 
Method & ScheduleFree+\tabularnewline
\hline 
\hline 
Parameters & 2B\tabularnewline
\hline 
Tokens & 40B\tabularnewline
\hline 
Batch-size & 1M\tabularnewline
\hline 
Sequence Length & 1024\tabularnewline
\hline 
GPUs & 32\tabularnewline
\hline 
Steps & 38147\tabularnewline
\hline 
Warmup & 2000\tabularnewline
\hline 
C-Warmup & 2000\tabularnewline
\hline 
Warm-started & No\tabularnewline
\hline 
Learning Rate & Polyak\tabularnewline
\hline 
Decay & 50\tabularnewline
\hline 
Clipping & 1\tabularnewline
\hline 
SF Beta & 0.9\tabularnewline
\hline 
AdamW Beta1 & 0.9\tabularnewline
\hline 
\end{tabular}

\subsection{Figure~\ref{fig:polyak}}
\begin{tabular}{|c|c|c|}
\hline 
Method & ScheduleFree Polyak & ScheduleFree 1/L1\tabularnewline
\hline 
\hline 
Parameters & 120M & 120M\tabularnewline
\hline 
Tokens & 10.5B & 10.5B\tabularnewline
\hline 
Batch-size & 0.5M & 0.5M\tabularnewline
\hline 
Sequence Length & 1024 & 1024\tabularnewline
\hline 
GPUs & 8 & 8\tabularnewline
\hline 
Steps & 20,000 & 20,000\tabularnewline
\hline 
Warmup & 781 & 781\tabularnewline
\hline 
C-Warmup & 781 & 781\tabularnewline
\hline 
Warm-started & No & No\tabularnewline
\hline 
Learning Rate & Polyak & Varies\tabularnewline
\hline 
Decay & 5 & 5\tabularnewline
\hline 
Clipping & 1 & 1\tabularnewline
\hline 
SF Beta & 0.9 & 0.9\tabularnewline
\hline 
AdamW Beta1 & 0.9 & 0.9\tabularnewline
\hline 
\end{tabular}

\subsection{Figure~\ref{fig:c_warmup}}
\begin{tabular}{|c|c|}
\hline 
Method & ScheduleFree+\tabularnewline
\hline 
\hline 
Parameters & 250M\tabularnewline
\hline 
Tokens & 5B\tabularnewline
\hline 
Batch-size & 1M\tabularnewline
\hline 
Sequence Length & 1024\tabularnewline
\hline 
GPUs & 8\tabularnewline
\hline 
Steps & 4769\tabularnewline
\hline 
Warmup & 400\tabularnewline
\hline 
C-Warmup & Varies\tabularnewline
\hline 
Warm-started & No\tabularnewline
\hline 
Learning Rate & Polyak\tabularnewline
\hline 
Decay & 5\tabularnewline
\hline 
Clipping & 1\tabularnewline
\hline 
SF Beta & 0.9\tabularnewline
\hline 
AdamW Beta1 & 0.9\tabularnewline
\hline 
\end{tabular}

\subsection{Figure~\ref{fig:sf_beta1}}

\begin{tabular}{|c|c|}
\hline 
Method & ScheduleFree+\tabularnewline
\hline 
\hline 
Parameters & 120M\tabularnewline
\hline 
Tokens & 200B\tabularnewline
\hline 
Batch-size & 0.5M\tabularnewline
\hline 
Sequence Length & 1024\tabularnewline
\hline 
GPUs & 8\tabularnewline
\hline 
Steps & 400,000\tabularnewline
\hline 
Warmup & 0\tabularnewline
\hline 
C-Warmup & 0\tabularnewline
\hline 
Warm-started & Yes\tabularnewline
\hline 
Learning Rate & Polyak\tabularnewline
\hline 
Decay & 5\tabularnewline
\hline 
Beta2 & 0.98\tabularnewline
\hline 
Clipping & 10\tabularnewline
\hline 
SF Beta & Varies\tabularnewline
\hline 
AdamW Beta1 & 0.9\tabularnewline
\hline 
r & 1\tabularnewline
\hline 
\end{tabular}

\subsection{Figure~\ref{fig:anneal_r}}
Configuration matches previous figure with SF beta annealed from 0.9 to 0.965.

\subsection{Figure~\ref{fig:curve-fit} a}

Fit parameters $a = 41.5$, $b = 3076$, $c = 1.88$

\begin{tabular}{|c|c|}
\hline 
Method & ScheduleFree+\tabularnewline
\hline 
\hline 
Parameters & 2B\tabularnewline
\hline 
Tokens & 40B\tabularnewline
\hline 
Batch-size & 1M\tabularnewline
\hline 
Sequence Length & 1024\tabularnewline
\hline 
GPUs & 32\tabularnewline
\hline 
Steps & 38147\tabularnewline
\hline 
Warmup & 400\tabularnewline
\hline 
C-Warmup & 400\tabularnewline
\hline 
Warm-started & No\tabularnewline
\hline 
Learning Rate & Polyak\tabularnewline
\hline 
Decay & 50\tabularnewline
\hline 
Clipping & 1\tabularnewline
\hline 
SF Beta & 0.9\tabularnewline
\hline 
AdamW Beta1 & 0.9\tabularnewline
\hline 
r & 0\tabularnewline
\hline 
\end{tabular}

\subsection{Figure~\ref{fig:curve-fit} b}

Fit parameters $a = 20.3$, $b = 377$, $c = 2.07$

\begin{tabular}{|c|c|}
\hline 
Method & ScheduleFree+\tabularnewline
\hline 
\hline 
Parameters & 500M\tabularnewline
\hline 
Tokens & 500B\tabularnewline
\hline 
Batch-size & 4M\tabularnewline
\hline 
Sequence Length & 1024\tabularnewline
\hline 
GPUs & 128\tabularnewline
\hline 
Steps & 119,210\tabularnewline
\hline 
Warmup & 400\tabularnewline
\hline 
C-Warmup & 400\tabularnewline
\hline 
Warm-started & No\tabularnewline
\hline 
Learning Rate & Polyak\tabularnewline
\hline 
Decay & 20\tabularnewline
\hline 
Clipping & 1\tabularnewline
\hline 
SF Beta & 0.9\tabularnewline
\hline 
AdamW Beta1 & 0.9\tabularnewline
\hline 
r & 1\tabularnewline
\hline 
\end{tabular}

\end{document}